\documentclass[11pt]{article}

\usepackage[final]{acl}
\usepackage{times}
\usepackage{kotex}
\usepackage{latexsym}
\usepackage{amssymb}
\usepackage{amsmath}
\usepackage{multirow}
\usepackage{booktabs}
\usepackage{array}
\usepackage{xurl}
\usepackage{tcolorbox}
\usepackage[T1]{fontenc}
\usepackage[utf8]{inputenc}

\usepackage{microtype}

\usepackage{inconsolata}
\usepackage{graphicx}

\title{EgoVoice: Proactive Spoken Assistance from Egocentric Multimodal Streams}

\author{Heeseung Kim \\
  Department of AI, University of Seoul \\
  \texttt{gmltmd789@uos.ac.kr} \\}

\begin{document}
\maketitle
\begin{abstract}
Wearable augmented reality (AR) assistants are moving toward continuous real-world interaction, where they perceive the user's activity through first-person video and audio and provide timely spoken guidance without being explicitly asked. While proactive video assistants, spoken dialog systems, and egocentric task understanding have each advanced rapidly, existing systems do not address the joint problem of deciding when to speak and what to say from continuous first-person streams. We introduce EgoVoice, a framework for training and evaluating proactive egocentric spoken assistants. From HoloAssist video recordings of real human instructors, we construct clean audio streams through source separation and speech resynthesis, and convert each video session into a format where the model must decide at each moment whether to remain silent or provide spoken guidance. We fine-tune an omni-modal LLM with our data, and further improve its proactive intervention behavior with direct preference optimization. Experiments across closed and open-source models show that existing systems rarely produce well-timed, meaningful proactive interventions, while EgoVoice yields clear improvements in intervention timing, content relevance, and human preference over the zero-shot backbone.\footnote{Project page: \url{https://egocentricvoice.github.io/}}
\end{abstract}

\section{Introduction}

As wearable devices such as smart glasses become more capable, augmented reality (AR) assistants are emerging as a practical interface for hands-free interaction. Meta Ray-Ban smart glasses combine voice interaction with first-person visual context, and research prototypes like Google Project Astra explore richer multimodal reasoning from egocentric camera feeds. The long-term vision goes further: an always-on assistant that perceives the user's ongoing activity through first-person (egocentric) video and audio, and provides spoken guidance at the right moment without being explicitly asked.

Despite growing interest in this direction, most voice interaction systems are designed and evaluated as reactive assistants. The user explicitly asks a question or gives a command, and the model generates a spoken or textual response~\cite{long2026vitaaudio, xie2024miniomnilanguagemodelshear, zeng2024glm4voiceintelligenthumanlikeendtoend}. In practice, however, real-world assistance often requires help before the user asks for it. During procedural tasks such as assembling furniture or following a recipe, users may be occupied, uncertain, or unaware that they need guidance. A user may not know that a mistake is about to occur or may be unable to pause and ask for help.  This motivates \textit{proactive egocentric spoken assistance}, where an assistant observes a first-person video and audio stream and decides whether to speak, when to speak, and what to say.

\begin{figure*}[t]
    \centering
    \includegraphics[width=0.93\textwidth]{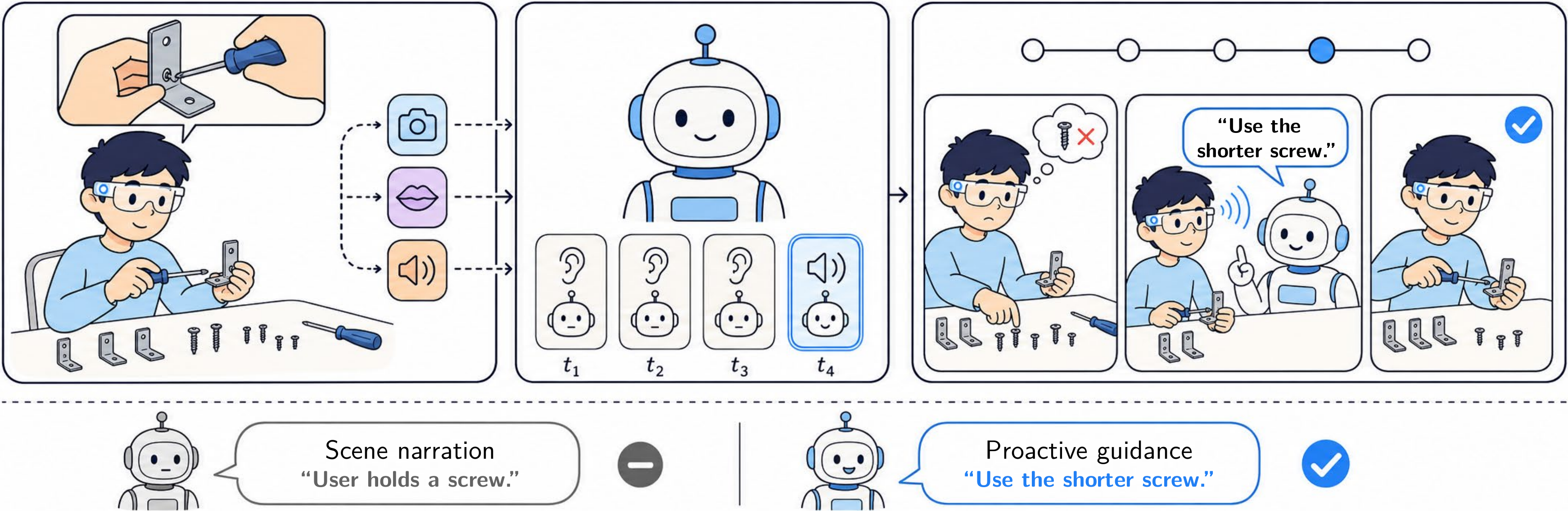}
    \caption{Overview of proactive egocentric spoken assistance. An AR assistant receives first-person video and audio (left), decides whether to remain silent or intervene (center), and provides spoken guidance (right).}
    \label{fig:1_overview}
    \vskip -14pt
\end{figure*}

Building such an assistant requires three capabilities that have so far been studied separately: multimodal perception with speech output, simultaneous listening and speaking, and proactive intervention from streaming egocentric video. Recent multimodal models can process video, audio, and text while generating spoken responses in real-time~\cite{fu2026vita,openai2024gpt4ocard,xu2025qwen3omnitechnicalreport,cui2026minicpmo45realtimefullduplex}. Full-duplex spoken dialog systems allow models to listen and speak simultaneously without explicit turn boundaries~\cite{defossez2024moshispeechtextfoundationmodel, roy2026personaplexvoicerolecontrol, wang2025freezeomni, yu2026salmonnomni}. Proactive video assistants can decide when and what guidance to offer from streaming egocentric video, though they produce text rather than speech~\cite{zhang-etal-2025-proactive, wang2026mmduet, Chen_2024_CVPR, Qian_2025_CVPR, zhang2026eyes}. However, to the best of our knowledge, no existing work combines these three capabilities to train a single model for proactive spoken guidance from continuous egocentric streams.

To address this gap, we introduce EgoVoice, a framework for training and evaluating proactive egocentric spoken assistants. We start from HoloAssist~\cite{Wang_2023_ICCV}, a corpus of egocentric task assistance with real human instructors, and process its recordings into clean, time-aligned streaming supervision through source separation and speech resynthesis. We then perform supervised fine-tuning (SFT) on MiniCPM-o 4.5~\cite{cui2026minicpmo45realtimefullduplex}, an open-source full-duplex multimodal LLM, to learn when and what to say from continuous egocentric video and audio streams, and further refine intervention timing and frequency through direct preference optimization (DPO)~\cite{rafailov2023direct}. We evaluate models along three axes: whether they intervene at the right time, whether the content is relevant, and whether the overall assistance is helpful as judged by a large language model (LLM) and by human evaluators.

Through experiments across closed-source, open-source, and our own trained models, we show that existing multimodal systems rarely produce well-timed proactive interventions, while EgoVoice yields improvements over all baselines on timing and content relevance, with complementary strengths between the SFT and DPO variants. Our contributions are as follows:

\begin{itemize}
    \item We construct a dataset and benchmark for proactive egocentric spoken assistance from HoloAssist recordings, producing clean training data with time-aligned intervention labels and spoken guidance targets.
    
    \item We train EgoVoice-SFT and EgoVoice-DPO, and show that the two stages are complementary: SFT retains an edge in content relevance while DPO refines intervention timing.
    
    \item We establish an evaluation protocol covering intervention timing, content relevance, and overall assistance quality for spoken assistant scenarios, and provide a comprehensive comparison across baselines and our models.
\end{itemize}
\section{Related Work}

\textbf{Spoken Dialog and Omni-Modal Models.} Early voice interaction systems relied on cascaded pipelines of automatic speech recognition (ASR), language modeling, and text-to-speech (TTS)~\cite{lin-etal-2024-advancing}. Subsequent work showed that LLMs can directly generate discrete speech tokens~\cite{long2026vitaaudio, xie2024miniomnilanguagemodelshear, zeng2024glm4voiceintelligenthumanlikeendtoend, zhang-etal-2023-speechgpt, kim2024paralinguisticsaware, fang-etal-2025-llama}. This line has rapidly scaled to omni-modal models that accept interleaved visual data, audio, and text while producing both text and speech in a streaming manner~\cite{fu2026vita, openai2024gpt4ocard, xu2025qwen3omnitechnicalreport, cui2026minicpmo45realtimefullduplex, xie2024miniomni2opensourcegpt4ovision, Zhong_2025_ICCV}.

A separate but related direction pursues full-duplex interaction, where models listen and speak simultaneously without explicit turn boundaries~~\cite{defossez2024moshispeechtextfoundationmodel, roy2026personaplexvoicerolecontrol, wang2025freezeomni, yu2026salmonnomni, zhang-etal-2025-omniflatten}. Among these, MiniCPM-o 4.5~\cite{cui2026minicpmo45realtimefullduplex} is especially close in modality coverage, as it supports full-duplex interaction with video and audio input, but it is designed as a general-purpose assistant rather than for egocentric task guidance. Despite this progress, these models are designed for reactive dialog and do not address proactive intervention during ongoing physical tasks.

\begin{figure*}[t]
\centering
\includegraphics[width=0.93\textwidth]{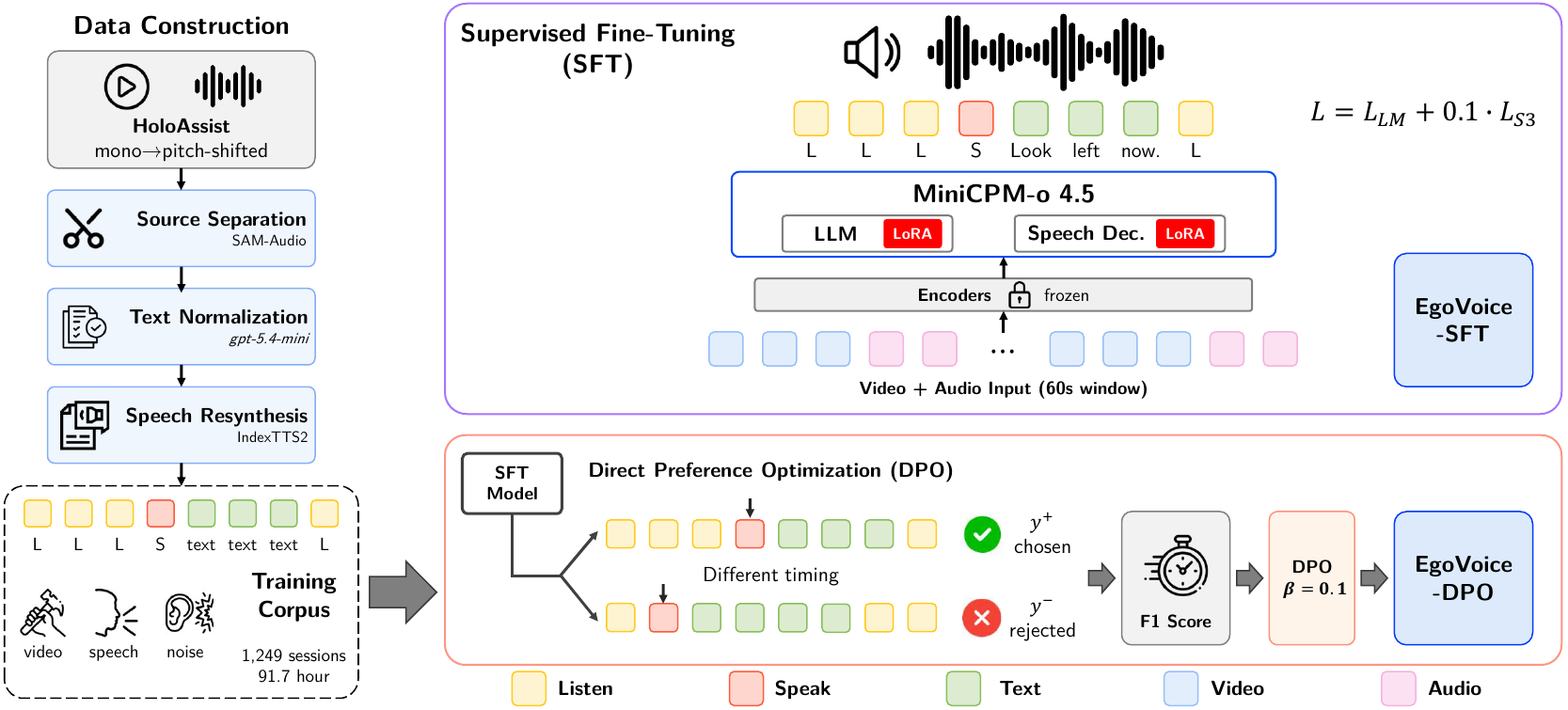}
\caption{Overview of the EgoVoice framework. We process HoloAssist recordings into clean training data (left, Section~\ref{sec:construction}), fine-tune MiniCPM-o 4.5 to learn listen-or-speak behavior (top right, EgoVoice-SFT), then refine intervention timing with DPO (bottom right, EgoVoice-DPO, Section~\ref{sec:training}).}
\label{fig:method}
\vskip -14pt
\end{figure*}

\textbf{Egocentric Task Understanding and Datasets.} Large-scale egocentric video datasets have enabled research on activity understanding from a first-person perspective. Ego4D~\cite{Grauman_2022_CVPR} established benchmarks for episodic memory, forecasting, and social interaction, and EPIC-KITCHENS~\cite{Damen_2018_ECCV} provided densely annotated cooking sessions for action recognition and anticipation. HoloAssist~\cite{Wang_2023_ICCV} captured instructor-guided task assistance sessions with synchronized egocentric video and audio, including annotations for intervention type prediction. Together with Assembly101~\cite{Sener_2022_CVPR} and Ego-Exo4D~\cite{Grauman_2024_CVPR}, these datasets have advanced procedure understanding, but none targets continuous spoken guidance during ongoing tasks. Among these, HoloAssist is closest to our setting as it contains real instructor speech, though its audio is recorded as a mono channel with pitch-shifted speaker identities, requiring additional processing before being used for voice interaction training.

\textbf{Proactive Assistance in Streaming Video.} Proactive assistance has been studied broadly in dialog systems, where models anticipate user needs and initiate responses without explicit prompts~\cite{ijcai2023p738, wu-etal-2019-proactive}. This idea has been extended to streaming video LLMs, which introduce per-frame speak-or-stay-silent decisions~\cite{Chen_2024_CVPR, wang-etal-2025-videollm} and improve timing through reinforcement learning, specialized objectives, or modular architectures~\cite{wang2026mmduet, Qian_2025_CVPR, lu2026auraalwaysonunderstandingrealtime, yan2026proactvlproactivevideollmrealtime}.

In egocentric settings, ProAssist~\cite{zhang-etal-2025-proactive} proposed a data curation pipeline, evaluation protocols, and an end-to-end streaming model for proactive dialog from video, and ESTP~\cite{zhang2026eyes} introduced a multi-stage training strategy for egocentric streaming video understanding. Prior work has shown that LLMs can generate appropriate content but struggle with the timing of spoken responses~\cite{umair-etal-2024-large}, and EgoSpeak~\cite{kim-etal-2025-egospeak} studied this problem by training an online prediction model on egocentric visual cues. Mirai~\cite{10.1145/3706599.3719881} and ProMemAssist~\cite{10.1145/3746059.3747770} demonstrated system prototypes for proactive spoken assistance on wearable devices, though they rely on cascaded API calls. These methods produce text, predict only when to speak, or rely on cascaded pipelines, rather than training a single model for intervention timing and spoken guidance from egocentric streams.

\section{Method}

EgoVoice converts egocentric instructional recordings into a format suitable for proactive spoken assistants. Building upon MiniCPM-o 4.5~\cite{cui2026minicpmo45realtimefullduplex}, a streaming full-duplex multimodal model, we adapt its listen-or-speak interface to egocentric task assistance through supervised fine-tuning and direct preference optimization. We first describe how we construct the dataset from HoloAssist (Section~\ref{sec:construction}), then present the evaluation protocol for measuring proactive spoken assistance (Section~\ref{sec:evaluation}), and finally the training pipeline that learns and refines intervention behavior (Section~\ref{sec:training}). An overview of EgoVoice is in Figure~\ref{fig:method}.

\subsection{Data Construction}
\label{sec:construction}

Perhaps the most straightforward approach to building an egocentric proactive spoken assistant would be to start from an existing video-text dataset~\cite{zhang-etal-2025-proactive, wang-etal-2025-videollm} and convert the textual instructions into speech. However, this strategy has fundamental limitations as illustrated in Figure~\ref{fig:tts_problem}. Text-based benchmarks assume that the full instruction is visible to the user at the moment it is issued. Speech, by contrast, takes time to play back. If a synthesized utterance is placed at the same timestamp as the original text instruction, the user will already be acting on the advice before the utterance finishes playing, creating a mismatch between the audio timeline and the user's behavior. Shifting the synthesis earlier to avoid mismatch is also unreliable, as the user's actions are continuous and the shifted utterance may collide with other instructions or fall during an unrelated action. Moreover, synthesizing speech from text labels does not account for the ambient sounds that naturally accompany real-world physical tasks.

For these reasons, we start from HoloAssist~\cite{Wang_2023_ICCV}, a large-scale corpus of egocentric task assistance in which a user performs physical tasks while receiving spoken guidance from a human instructor. HoloAssist's spoken interventions are already aligned with the user's behavior through real-world recording, and preserve background noises. However, the raw recordings are not directly usable for training a multimodal voice interaction model. The video comes with a single audio channel in which user speech, instructor speech, and background sounds are mixed, and speaker identities are anonymized through pitch shifting, which makes them sound unnatural. We therefore process each session as follows.

First, we separate background audio from the mixture. The pitch shifting renders the voices unsuitable for training a natural-sounding voice interaction model, so we isolate only the background sounds. We use SAM-Audio~\cite{shi2025samaudiosegmentaudio}, a prompt-based source separation model, to extract human voices from the mixture using the speech-containing intervals as prompts and retain the residual as the background channel. We then apply Silero VAD~\cite{SileroVAD} to verify that no residual human speech remains in the separated background, and discard all sessions where voice activity is still detected.

Second, we synthesize clean user and instructor speech. The original transcripts in the HoloAssist metadata contain spelling errors, inconsistent punctuation, and disfluencies, which we normalize using \textit{gpt-5.4-mini}~\cite{singh2026openaigpt5card}. We also remove task-irrelevant opening and closing remarks (9.0\% of utterances). We then resynthesize with IndexTTS2~\cite{Zhou_Zhou_He_Zhou_Wang_Deng_Shu_2026}, a speaker-adaptive TTS, with per-session reference speakers sampled from the English subset of Multilingual LibriSpeech~\cite{pratap20_interspeech} (2,498 training speakers, 340 disjoint evaluation speakers). Each utterance is synthesized multiple times; candidates exceeding the original duration are discarded to preserve timing, and the candidate whose ASR transcription exactly matches the original text, as verified by Qwen3-ASR-1.7B~\cite{shi2026qwen3asrtechnicalreport}, is selected. Sessions containing any mispronounced utterance are discarded after manual verification. Further details are provided in Appendix~\ref{app:data}.

\begin{figure*}[t]
\centering
\includegraphics[width=0.93\textwidth]{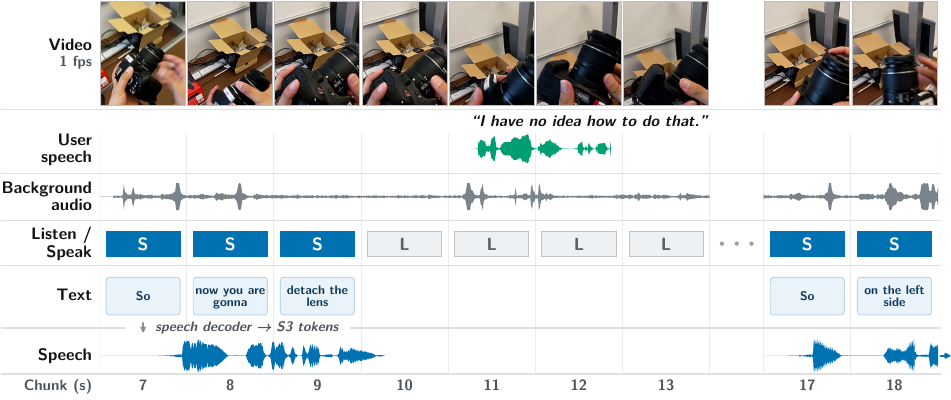}
\caption{The streaming interface of EgoVoice (\texttt{R101-2Aug-DSLR}). Each chunk carries one video frame and one second of audio, and the model emits \textsc{Listen} or \textsc{Speak} for each. On \textsc{Speak} chunks the LLM emits text tokens, and the speech-token decoder turns them into S3 speech tokens that are decoded into the waveform the user hears.}
\label{fig:stream}
\vskip -14pt
\end{figure*}
After this full pipeline, the original HoloAssist corpus of 1,673 instructional sessions (124.8 hours) yields 1,249 training sessions (91.7 hours) and 170 held-out benchmark sessions (12.9 hours). Each processed session contains synchronized egocentric video, resynthesized user speech, resynthesized instructor speech, and separated background audio. 

\subsection{Evaluation Protocol}
\label{sec:evaluation}

We propose an evaluation protocol for measuring proactive spoken assistance over full egocentric task sessions. Following ProAssist~\cite{zhang-etal-2025-proactive}, we provide each model with a system prompt containing the task name and step-by-step recipe, so that models have access to the procedural context when deciding whether and what to speak. 

\textbf{Onset Timing.} We match predicted utterances to reference instructor utterances using the Hungarian algorithm~\cite{https://doi.org/10.1002/nav.3800020109} with an asymmetric onset window of $-2.5$s (early) to $+1.5$s (late), reflecting that early interventions are less disruptive than late ones. A predicted utterance can match a reference only if its onset falls within this window. From the matched pairs, we compute Precision, Recall, and F1 as count-based scores over the number of matched, predicted, and reference utterances. All timing metrics are computed per session and macro-averaged across the 170 benchmark sessions.

\textbf{End-of-Utterance Timing.} Onset metrics do not capture whether the model finishes speaking before the user needs to act. We therefore measure end-vs-action, the gap between the start of the user's next action and the predicted utterance end time. A positive value means the model finished speaking before the user needed to act; a negative value means the model was still talking when the user began the action. Since this metric depends on when an utterance begins and how long it lasts, we also report its decomposition. We define start-vs-action as the gap between the start of the user's next action and the predicted utterance start time; end-vs-action then equals start-vs-action minus the utterance duration for every matched pair (Appendix~\ref{app:eval}).

Since end-vs-action is meaningful only when the utterance is a procedural directive that prompts a user action rather than a brief acknowledgement, we define an action-linked evaluation subset. Using \textit{gpt-5.5}~\cite{singh2026openaigpt5card}, we classify each reference utterance as a procedural directive or non-directive and link directives to the user's nearest subsequent action within $T{=}5$ seconds, yielding 942 of 2,067 reference utterances across 170 benchmark sessions. Unlike the other metrics, end-vs-action is reported only on this action-linked subset.

\textbf{Content Relevance.} For each temporally matched pair, we compute cosine similarity between the predicted and reference texts using Qwen3-Embedding-0.6B~\cite{zhang2025qwen3embeddingadvancingtext}. 

\textbf{Overall Assistance Quality.} We use an LLM-as-a-judge evaluation following the session-level assessment protocol of ProAssist~\cite{zhang-etal-2025-proactive}. We prompt \textit{gpt-5.4-mini}~\cite{singh2026openaigpt5card} (reasoning effort=medium) to rate each session on four axes on a 1-to-5 Likert scale: Correctness (whether each utterance is accurate given the reference), Promptness (whether the assistant intervenes at the right moment), Efficiency (whether responses are concise and free of redundancy), and Overall (the overall helpfulness of the assistant's behavior). The judge receives the full session context, with reference and predicted utterances timestamped. Further details are provided in Appendix~\ref{app:eval}.

\textbf{Evaluating Intended and Delivered Speech.} Most multimodal speech models produce both a text response and a spoken response. We therefore evaluate all metrics from two sources: \texttt{text-llm}, the text that the model generates internally (reported in Appendix~\ref{app:textllm}), and \texttt{text-asr}, the text obtained by transcribing the model's spoken output with ASR. The former reflects what the model intended to say; the latter reflects what the user actually hears. For timing, \texttt{text-llm} uses the chunk index at which the model produces the text (described in Section~\ref{sec:training}). For \texttt{text-asr}, we first transcribe the generated audio with Qwen3-ASR-1.7B, then obtain word-level alignment using Qwen3-ForcedAligner-0.6B~\cite{shi2026qwen3asrtechnicalreport} to determine the precise onset and offset of each utterance.

\subsection{Training}
\label{sec:training}

We train EgoVoice by fine-tuning MiniCPM-o 4.5~\cite{cui2026minicpmo45realtimefullduplex}, a streaming full-duplex multimodal model that processes interleaved video and audio chunk by chunk. At each one-second chunk, the model encodes the current video frame and audio segment (background audio and user speech), appends the resulting tokens to a running context, and produces a decision: either a \textsc{Listen} token, indicating that the model should remain silent, or a \textsc{Speak} token, after which the LLM backbone autoregressively generates a text response. For each text token, the corresponding LLM hidden state is passed to a speech-token decoder (0.3B parameters), which produces S3 speech tokens~\cite{du2024cosyvoicescalablemultilingualzeroshot} converted to a waveform by a streaming flow-matching decoder~\cite{lipman2023flow}. Figure~\ref{fig:stream} shows this process on a benchmark session.

For training, we split each processed session from Section~\ref{sec:construction} into 60-second windows at one-second chunk granularity. Each chunk $t$ within a window is represented as $s_t = (v_t, a_t, z_t, w_t, c_t)$, where $v_t$ is the egocentric video frame, $a_t$ is the input audio containing background audio and user speech, $z_t \in \{\textsc{Listen}, \textsc{Speak}\}$ is the intervention label, $w_t$ is the normalized text instruction, and $c_t$ is the corresponding S3 speech-token sequence (both empty when $z_t = \textsc{Listen}$). Instructor speech is never included in the model input; it is used exclusively as the target output.

Each instructor utterance is segmented into one-second chunks using word-level timestamps from Qwen3-ForcedAligner-0.6B, distributing the text and speech tokens across the corresponding chunks. For each reference utterance, we construct a 60-second training window centered on that utterance (with slight temporal jitter), so that every window contains at least one intervention event surrounded by listening context. This yields 13,676 training windows per epoch. At evaluation, models process complete sessions without windowing.

\textbf{Supervised Fine-Tuning (SFT).} We first train the model to learn the basic behavior of proactive spoken assistance: remain silent during most chunks, but generate task-relevant spoken guidance when intervention is needed. We apply LoRA~\cite{hu2022lora} to both the LLM backbone and the speech-token decoder, while keeping the vision encoder, audio encoder, and other adapters frozen. Within each window, the model processes chunks autoregressively, emitting either a \textsc{Listen} token or a \textsc{Speak} token followed by the guidance text in byte-pair encoding (BPE) tokens and the S3 speech-token sequence. The training loss is $\mathcal{L}_{\mathrm{SFT}} = \mathcal{L}_{\mathrm{LM}} + \lambda \cdot \mathcal{L}_{\mathrm{S3}}$ with $\lambda = 0.1$, where $\mathcal{L}_{\mathrm{LM}}$ is a single autoregressive cross-entropy over \textsc{Listen}/\textsc{Speak} decisions, BPE text tokens, and end-of-turn markers, and $\mathcal{L}_{\mathrm{S3}}$ is a cross-entropy on the speech-token decoder over S3 speech-token frames for \textsc{Speak} chunks only. Gradients from $\mathcal{L}_{\mathrm{S3}}$ are stopped at the LLM hidden states, so the two losses do not share backward signal through the backbone. 

\textbf{Direct Preference Optimization (DPO).} Supervised fine-tuning provides a strong starting point, but proactive assistance admits multiple acceptable behaviors. We further refine the SFT model with DPO~\cite{rafailov2023direct}, allowing the model to learn from comparisons between alternative candidate samples under the same streaming context.

From the SFT model, we generate $N{=}10$ candidate samples per training window, each conditioned on the same visual and audio context. We score each candidate with the F1 defined in Section~\ref{sec:evaluation}, computed after temporal matching using \texttt{text-llm}. We construct preference pairs by selecting the candidate with the highest F1 as the chosen sample $y^+$ and randomly sampling the rejected sample $y^-$ from the bottom three candidates by F1, requiring a minimum F1 gap of 0.10. This yields 5,335 preference pairs from windows across 1,021 training sessions.

We train with length-normalized DPO~\cite{lambert2025tulu}. Given a streaming context $x$ and a preference pair $(y^+, y^-)$, the objective is:
\begin{equation}
\mathcal{L}_{\mathrm{DPO}} = -\mathbb{E}\!\left[\log \sigma\!\left(\frac{\beta}{|y^+|} \rho_{y^+} - \frac{\beta}{|y^-|} \rho_{y^-}\right)\right],
\label{eq:dpo}
\end{equation}
where $\rho_y = \log \pi_\theta(y \mid x) - \log \pi_{\mathrm{ref}}(y \mid x)$ and $\beta$ is a hyperparameter controlling the strength of the preference constraint. Here, $y$ denotes the sequence of \textsc{Listen}/\textsc{Speak} decisions and generated text tokens over the window; S3 speech tokens are excluded from the DPO likelihood. The reference model $\pi_{\mathrm{ref}}$ is the frozen SFT model. For DPO, we apply LoRA to the LLM backbone only, since preferences are constructed over intervention decisions and textual guidance rather than speech synthesis.
\section{Experiments}
\label{sec:experiments}

\begin{table*}[t]
\centering
\small
\begin{tabular}{l ccc c cccc c}
\toprule
\multirow{2}{*}{\textbf{Model}} & \multicolumn{3}{c}{\textbf{Onset Timing}} & \textbf{Content} & \multicolumn{4}{c}{\textbf{LLM Judge (1--5)}} & \textbf{End Timing} \\
\cmidrule(lr){2-4} \cmidrule(lr){5-5} \cmidrule(lr){6-9} \cmidrule(lr){10-10}
& \textbf{Prec.} & \textbf{Rec.} & \textbf{F1} & \textbf{Cosine} & \textbf{Correct.} & \textbf{Prompt.} & \textbf{Effic.} & \textbf{Overall} & \textbf{end-vs-act (s)} \\
\midrule
Gemini Live & .379 & .019 & .022 & .448 & 2.09 & 2.39 & 2.67 & 1.94 & $-$3.20 / $-$2.26 \\
\texttt{gpt-realtime} & .285 & .024 & .030 & .500 & 2.14 & 2.60 & 2.52 & 2.10 & $-$7.40 / $-$7.46 \\
\texttt{gpt-realtime-2} & .256 & .019 & .024 & .456 & 2.28 & 2.40 & 2.10 & 2.14 & $-$16.1 / $-$16.8 \\
\midrule
ProAssist$^\dagger$ & \underline{.480} & .158 & .195 & .594 & 2.34 & \underline{2.63} & \textbf{2.99} & 2.06 & -- \\
\quad + ASR/TTS cascade & .479 & .158 & .195 & .564 & 2.13 & 2.50 & \underline{2.87} & 1.92 & $-$0.44 / $-$1.56 \\
MiniCPM-o 4.5 & .259 & .108 & .113 & .519 & 1.54 & 1.63 & 1.73 & 1.41 & $-$4.69 / $-$0.40 \\
\midrule
EgoVoice-SFT & .467 & \underline{.235} & \underline{.274} & \textbf{.684} & \textbf{2.41} & \textbf{2.64} & 2.64 & \textbf{2.23} & \underline{+1.57 / +1.52} \\
EgoVoice-DPO & \textbf{.482} & \textbf{.285} & \textbf{.324} & \underline{.649} & \underline{2.35} & 2.55 & 2.42 & \underline{2.18} & \textbf{+1.75 / +1.62} \\
\midrule
Human instructor$^\ddagger$ & -- & -- & -- & -- & -- & -- & -- & -- & +1.69 / +1.38 \\
\bottomrule
\end{tabular}
\caption{Results on 170 benchmark sessions (\texttt{text-asr}). $\dagger$~indicates a text-only baseline, evaluated on \texttt{text-llm}. $\ddagger$~indicates the human instructor, whose utterances are the reference for all systems. End-vs-act is reported as mean / median on the action-linked subset ($T{=}5$), where positive values mean the model typically finishes speaking before the user acts; per-intervention distributions are in Appendix~\ref{app:enddist}. Best in \textbf{bold}, second best \underline{underlined}.}
\label{tab:main}
\vskip -14pt
\end{table*}

\textbf{Baselines.} We compare EgoVoice against six baselines spanning three categories. For closed-source APIs, we evaluate Gemini Live (\texttt{gemini-live-2.5-flash-native-audio}) and two snapshots of the OpenAI Realtime API (\texttt{gpt-realtime} and \texttt{gpt-realtime-2}), all of which accept streaming video and audio and produce both text and speech. Since these APIs rely on voice activity detection and rarely respond without active user speech, we remove background audio to prevent false triggering, providing a more favorable evaluation condition.

As a reference open-source text baseline, we evaluate ProAssist~\cite{zhang-etal-2025-proactive} (8B), the most directly comparable prior work, which uses the HoloAssist data and evaluation protocol but handles text rather than speech. Following the ProAssist protocol, it receives streaming video together with ground-truth transcripts of the user's speech at their original timestamps, and a neutral opening query (e.g., ``help me assemble this lamp'') for sessions in which the user never speaks.

We also evaluate two open-source baselines that support voice interaction. The cascaded baseline (+ ASR/TTS cascade in Table~\ref{tab:main}) pairs ProAssist with Qwen3-ASR-1.7B~\cite{shi2026qwen3asrtechnicalreport} for the user's speech and IndexTTS2~\cite{Zhou_Zhou_He_Zhou_Wang_Deng_Shu_2026} for the assistant's speech, using the same per-session reference speaker as our benchmark. We assume ideal zero latency between ASR, ProAssist, and TTS, so that each utterance begins playing at the chunk where ProAssist generates it and only the duration of the synthesized speech affects when it ends. The end-to-end baseline is MiniCPM-o 4.5~\cite{cui2026minicpmo45realtimefullduplex} (9B) in a zero-shot setting, which also serves as EgoVoice's backbone before fine-tuning.

\textbf{Our Models.} We evaluate EgoVoice-SFT, trained with SFT on our processed data, and EgoVoice-DPO, which further refines EgoVoice-SFT with DPO on intervention timing. 

\textbf{Implementation Details.} For SFT, we apply LoRA ($r{=}32$, $\alpha{=}64$) to both the LLM backbone and the speech-token decoder, training for 5 epochs on 13,676 windows with a learning rate of 2e-4 and a batch size of 32. For DPO, we train EgoVoice-SFT with $\beta{=}0.1$ for 1 epoch with a learning rate of 5e-6 and a batch size of 16, and select the checkpoint based on validation performance. Additional details are provided in Appendices~\ref{app:baselines} and ~\ref{app:hparams}.
\section{Results}
\label{sec:results}

\begin{table*}[t]
\centering
\small
\begin{tabular}{l ccc c cccc c}
\toprule
\multirow{2}{*}{\textbf{Variant}} & \multicolumn{3}{c}{\textbf{Onset Timing}} & \textbf{Content} & \multicolumn{4}{c}{\textbf{LLM Judge (1--5)}} & \textbf{End Timing} \\
\cmidrule(lr){2-4} \cmidrule(lr){5-5} \cmidrule(lr){6-9} \cmidrule(lr){10-10}
& \textbf{Prec.} & \textbf{Rec.} & \textbf{F1} & \textbf{Cosine} & \textbf{Correct.} & \textbf{Prompt.} & \textbf{Effic.} & \textbf{Overall} & \textbf{end-vs-act (s)} \\
\midrule
\multicolumn{10}{l}{\textbf{\textit{SFT design choices}}} \\
EgoVoice-SFT & .467 & .235 & .274 & .684 & 2.41 & 2.64 & 2.64 & 2.23 & +1.57 / +1.52 \\
\quad w/o recipe & .398 & .223 & .240 & .621 & 1.91 & 2.14 & 2.15 & 1.81 & +2.32 / +2.09 \\
\quad w/o detach & .435 & .246 & .276 & .680 & 2.38 & 2.61 & 2.45 & 2.18 & +2.19 / +1.86 \\
\midrule
\multicolumn{10}{l}{\textbf{\textit{DPO variants (all from EgoVoice-SFT)}}} \\
EgoVoice-DPO & .482 & .285 & .324 & .649 & 2.35 & 2.55 & 2.42 & 2.18 & +1.75 / +1.62 \\
Vanilla DPO, $\beta$=0.1 & .458 & .261 & .298 & .651 & 2.37 & 2.70 & 2.58 & 2.25 & +1.14 / +1.40 \\
SimPO  & .457 & .309 & .328 & .641 & 2.38 & 2.60 & 2.40 & 2.18 & +1.86 / +1.77 \\
Length-norm., $\beta$=0.5 & .401 & .371 & .345 & .617 & 2.22 & 2.51 & 2.02 & 2.04 & +1.62 / +1.62 \\
F1 + Cosine reward & .451 & .205 & .248 & .674 & 2.36 & 2.61 & 2.60 & 2.16 & +1.51 / +1.53 \\
\bottomrule
\end{tabular}
\caption{Ablation results (\texttt{text-asr}). Metrics are computed following the same protocol as Table~\ref{tab:main}. EgoVoice-DPO uses length-normalized DPO with $\beta=0.1$. Each variant differs from this reference in only the listed factor.}
\label{tab:ablation}
\vskip -14pt
\end{table*}

We evaluate all models on the 170 benchmark sessions. Unless otherwise noted, we report metrics on the \texttt{text-asr} channel, which reflects what the user actually hears; \texttt{text-llm} results and qualitative examples are provided in Appendix~\ref{app:additional}. Reported \textit{p}-values are from a paired Wilcoxon signed-rank test~\cite{c4091bd3-d888-3152-8886-c284bf66a93a} over sessions.

\subsection{Model Comparison}
\label{sec:comparison}

Table~\ref{tab:main} shows that off-the-shelf models struggle to produce well-timed proactive interventions. The closed-source APIs all achieve F1 scores of at most 0.03, as they rely on server-side VAD that triggers responses only when the user speaks; since users are largely silent during physical tasks, these models rarely find an occasion to intervene. MiniCPM-o 4.5 in a zero-shot setting performs better (F1 = 0.113) by virtue of its listen-or-speak mechanism, but its judge scores are the lowest overall (Overall = 1.41), reflecting a tendency to narrate observed actions rather than provide task-relevant guidance. The same behavior shows up in its end-vs-action, where a tail of very long narrations pulls the mean far below the median ($-$4.69s against $-$0.40s). 

EgoVoice-SFT substantially improves over MiniCPM-o 4.5 across all metrics, with \textit{p} < $10^{-12}$ on both onset F1 and LLM Judge (Overall). EgoVoice-SFT also outperforms ProAssist on F1 score (0.274 vs 0.195, \textit{p} < $10^{-4}$) and cosine similarity (0.684 vs 0.594), even though ProAssist is a text-only model evaluated using \texttt{text-llm}. Nevertheless, ProAssist achieves the highest Efficiency score among all models. It speaks less often than EgoVoice and in longer turns (2.2 utterances of 16.6 words per session, against 5.6 utterances of 7.8 words), with almost no brief acknowledgements: utterances of three words or fewer make up 0.8\% of its output, against 19.3\% for EgoVoice-DPO and 17.9\% for the instructor. The Efficiency axis appears to favor this turn structure (Appendix~\ref{app:uttstats}).

Turning ProAssist into a spoken assistant (+ ASR/TTS cascade in Table~\ref{tab:main}) does not close the gap: because the cascade inherits ProAssist's intervention decisions, its onset F1 is unchanged at 0.195. In addition, the cascade's speech ends after the user has already begun the linked action in 62.0\% of matched interventions, against 24.3\% for EgoVoice-SFT and 28.2\% for EgoVoice-DPO, giving it negative mean and median end-vs-action values. Deciding when to speak is therefore not sufficient on its own, because the length of the resulting speech determines whether the guidance arrives before it is needed.

EgoVoice-SFT and EgoVoice-DPO are the only models with positive end-vs-action by both mean and median, and their margins (+1.57s and +1.75s) are close to the instructor's own (+1.69s). Between the two, EgoVoice-DPO further improves onset timing, raising F1 by 0.050 (\textit{p} = 0.0007) as both precision and recall increase, indicating that DPO encourages the model to intervene at more of the right moments rather than simply producing more speech. This comes at a small cost in content similarity (0.684 to 0.649), while LLM Judge (Overall) does not separate the two variants (95\% bootstrap CI $[-0.224, 0.072]$, \textit{p} = 0.42). The two stages are thus complementary: EgoVoice-DPO prioritizes intervention timing, while EgoVoice-SFT retains a small edge on content metrics.

\subsection{Ablation Studies}
\label{sec:ablations}

Table~\ref{tab:ablation} examines individual design choices. For SFT, removing the recipe from the system prompt so that the model relies solely on video and audio causes the largest degradation across all metrics except end-vs-action. This confirms that procedural context is essential for generating relevant guidance, not just for what to say but also for when to say it. We also examine the effect of gradient detachment between the LLM backbone and the speech-token decoder. Removing detachment yields slightly higher F1 but lower judge scores, suggesting that speech loss gradients flowing into the LLM backbone interfere with content quality.

For DPO, we first compare three loss variants: length-normalized DPO~\cite{lambert2025tulu}, which normalizes by sequence length; vanilla DPO~\cite{rafailov2023direct}; and SimPO~\cite{NEURIPS2024_e099c1c9}, a reference-free variant. All three improve F1 over SFT, with modest differences among them: trade-offs across cosine similarity and judge scores rather than clear wins. We select length-normalized DPO as our default because it provides the most balanced performance across timing, content, and end-vs-action metrics.

We then examine the sensitivity to $\beta$ and the reward strategy. Increasing $\beta$ beyond 0.10 leads the model to speak more often without meaningful content, which raises recall and superficially improves F1 but degrades precision and other metrics related to response quality. To address the content quality decline, we also construct preference pairs requiring improvement on both F1 and cosine similarity. However, this composite strategy instead leads to degradation in both metrics relative to SFT. Selecting by F1 alone proves more effective, suggesting that a single clear optimization signal is preferable to a composite one in this setting.

\subsection{Human Evaluation}
\label{sec:human_eval}

We additionally run pairwise human preference evaluations on Amazon Mechanical Turk. Each evaluator watches two versions of the same egocentric clip, differing only in which assistant's spoken response is overlaid in the clip, and selects the preferred assistant or a tie. We recruit 100 workers per comparison and discard those who fail a built-in attention check, retaining 252--267 valid judgments per comparison. The full protocol and the evaluation template are in Appendix~\ref{app:human_eval}.

As shown in Table~\ref{tab:human}, both EgoVoice-SFT and EgoVoice-DPO are preferred over MiniCPM-o 4.5 (\textit{p} < 0.05), confirming that the improvements observed in automatic metrics translate to perceptible gains in human judgment. Evaluators show a slight numerical preference for DPO over SFT, but the difference is not significant (\textit{p} = 0.60), suggesting that the timing gains from DPO and the content advantage of SFT are perceived as comparably valuable.

\subsection{Analysis}
\label{sec:analysis}

\begin{table}[t]
\centering
\small
\begin{tabular}{lcccc}
\toprule
\textbf{Comparison} & \textbf{Win} & \textbf{Tie} & \textbf{Lose} & \textbf{\textit{p}-val} \\
\midrule
SFT vs MiniCPM & \textbf{45.6\%} & 23.0\% & 31.3\% & 0.01 \\
DPO vs MiniCPM & \textbf{45.3\%} & 21.7\% & 33.0\% & 0.02 \\
DPO vs SFT & \textbf{36.1\%} & 30.6\% & 33.3\% & 0.60 \\
\bottomrule
\end{tabular}
\caption{Pairwise human preference evaluation results.}
\label{tab:human}
\vskip -14pt
\end{table}

\textbf{Intervention Ordering.} Onset F1 measures whether interventions land at the right moments, but not whether they follow the order of the task, since the Hungarian matching pairs each prediction with a reference independently of every other pair. A model could therefore match many references while explaining a later step before an earlier one. We test this with Kendall's $\tau_b$, a rank correlation that counts how often two orderings agree on which of a pair of items comes first, ranging from $-1$ for a reversed order to $1$ for an identical one, with a correction for tied ranks~\cite{df984cd0-5ddb-387f-ae81-1c7b41e92c35}. We compute it per session between the order in which the model produced its matched interventions and the order of the references they matched. Over sessions with at least two matched pairs, EgoVoice-DPO reaches 0.868 and EgoVoice-SFT 0.841, against 0.850 for ProAssist and 0.749 for MiniCPM-o 4.5. The trained models thus move through the task in close to the order the instructor did, while the zero-shot backbone is markedly less consistent.

\textbf{Content Alignment.} A model could deliver guidance in the right order yet phrase it generically enough to fit any nearby step, which onset matching alone would not detect. For each matched pair we therefore compare the prediction against its matched reference and the two references before and the two after it, using the same cosine similarity as our Content Relevance metric, and record which is closest. The matched reference is closest for 59.5\% of EgoVoice-SFT pairs and 52.8\% of EgoVoice-DPO, against 45.9\% for ProAssist and 31.8\% for MiniCPM-o 4.5. All systems thus carry some step-specific content, but the trained models are the most locally precise.

\begin{table}[t]
\centering
\small
\begin{tabular}{lccc}
\toprule
\textbf{Model} & \textbf{Guidance} & \textbf{Narration} & \textbf{Other} \\
\midrule
Human instructor & 76.7 & 5.6 & 17.8 \\
\midrule
MiniCPM-o 4.5 & 30.1 & 21.0 & 49.0 \\
\midrule
EgoVoice-SFT & 65.7 & 5.4 & 29.0 \\
EgoVoice-DPO & 64.2 & 5.5 & 30.3 \\
\bottomrule
\end{tabular}
\caption{Distribution of utterance types on the 170 benchmark sessions (\%). Each utterance is classified as proactive guidance, third-person scene narration, or other. The first row is the reference instructor speech.}
\label{tab:utttype}
\vskip -14pt
\end{table}

\textbf{Utterance Types.} Neither timing nor content metrics say what kind of speech a model produces. We classify every delivered utterance, and every reference utterance from the human instructors, into proactive guidance, third-person scene narration, and other using \textit{gpt-5.6-luna} (Table~\ref{tab:utttype}; the prompt and protocol are in Appendix~\ref{app:eval}). Our models deliver guidance more often than the backbone and bring narration close to the instructors' level. What remains is the other category, which for every system consists mostly of brief acknowledgments: 70.7\% of EgoVoice-DPO's are three words or fewer. EgoVoice-SFT and EgoVoice-DPO generate more of them than the instructors, and that gap accounts for most of the remaining distance to the reference distribution.

\textbf{Failure Modes.} Quantitative results in Table~\ref{tab:main} hide two failure modes. First, a positive mean end-vs-action does not guarantee the model finishes in time: EgoVoice-DPO still finishes after the user has begun the linked action in 28.2\% of matched interventions and EgoVoice-SFT in 24.3\%, which is well below MiniCPM-o 4.5 (54.5\%) and the cascade (62.0\%), but far from solved. Second, recall is the binding constraint: at a recall of 0.285, EgoVoice-DPO misses most moments where the instructor intervened. Appendix~\ref{app:qualitative} groups examples of each by failure type.

\section{Conclusion}

We presented EgoVoice, a framework for training and evaluating proactive egocentric spoken assistants. We constructed a spoken interaction dataset from HoloAssist through source separation and speech resynthesis, and proposed an evaluation protocol covering intervention timing, content relevance, and overall assistance quality. Through supervised fine-tuning and direct preference optimization on MiniCPM-o 4.5, we showed that both EgoVoice-SFT and EgoVoice-DPO outperform all baselines on onset F1 and content relevance. We hope that EgoVoice serves as an initial step toward building always-on spoken assistants that can proactively guide users through real-world tasks.
\section*{Limitations}
While we demonstrated the potential of egocentric proactive spoken assistance, our framework has several limitations. First, from a data perspective, our dataset is constructed from HoloAssist, which covers a specific set of procedural assembly tasks. As a result, the model's ability to generalize to other domains such as cooking, outdoor navigation, or social interaction is not guaranteed. In addition, there is still a substantial performance gap between our model and the ground truth instructor behavior (Appendix~\ref{app:enddist}), suggesting that more diverse training data and larger-scale training would be needed to approach human-level proactive assistance.

Second, our backbone model, MiniCPM-o 4.5, occasionally exhibits several issues that we observed during our experiments, including excessively long responses, severe repetition, and occasional language switching (e.g., generating Chinese tokens or unrecognizable special characters in the middle of English utterances). Although these artifacts are reduced through our fine-tuning, they still carry over into both EgoVoice-SFT and EgoVoice-DPO, as LoRA adapts only a small subset of model parameters. The model also inherits the one-second chunk granularity of MiniCPM-o 4.5, which may limit the precision of intervention timing in scenarios that require sub-second responsiveness.

Third, our evaluation is conducted on prerecorded sessions: the user's actions are already fixed, so every model is scored against behavior that is independent of its own advice. Our numbers should therefore be read as a proxy for interactive performance rather than a measurement of it. We also do not model overlap between the two speakers: each utterance plays to completion, so the protocol covers neither a user interrupting the assistant nor the assistant yielding mid-utterance. Handling barge-in in both directions is necessary for deployment and is left to future work.

Fourth, our DPO stage improves intervention timing but introduces a slight decline in content quality, indicating that the reward signal has room for further refinement. Finally, deploying an always-on spoken assistant in real-world settings raises privacy risks: continuous egocentric recording may capture bystanders without consent, and transmitting audio and video to external servers risks exposing sensitive information. Mitigating these risks will require bystander consent mechanisms, on-device processing to minimize data transmission, and audio redaction protocols. Nevertheless, with growing industry investment in AR and smart glasses, we believe open-source contributions toward proactive spoken assistance will become increasingly valuable, and we hope our dataset, code, and models will serve as a useful foundation for this emerging direction.
\section*{Ethics Statement}

Our dataset is derived from HoloAssist~\cite{Wang_2023_ICCV}, released under CDLAv2, which permits redistribution and derivative works, and we use all artifacts consistently with their intended use (Appendix~\ref{app:licenses}). The original recordings anonymize speaker identities by pitch shifting, and our resynthesis replaces every voice with zero-shot TTS output from reference speakers in Multilingual LibriSpeech~\cite{pratap20_interspeech} (CC BY 4.0), a corpus built from public-domain LibriVox audiobook recordings. Both corpora identify speakers only by numeric identifiers; the speech we release is entirely synthetic, and the separated background channel is verified to contain no residual speech (Section~\ref{sec:construction}). No voice in our data can therefore be linked to a named person, and we will preserve this property in any future version of the dataset.

For the human evaluation, we recruited workers on Amazon Mechanical Turk, compensated them for each completed task, and collected no personally identifying information beyond what the platform requires; the protocol is described in Appendix~\ref{app:human_eval}. Beyond the study itself, an always-on assistant that continuously records first-person video and audio carries risks for the user and for bystanders. We discuss these risks and possible mitigations in the Limitations section, and we release our models for research on proactive assistance rather than for deployment as a monitoring tool.
\section*{Acknowledgments}

This work was supported by the Advanced GPU Utilization Support Program funded by the Government of the Republic of Korea (Ministry of Science and ICT), by the National Research Foundation of Korea (NRF) grant funded by the Korea Government (MSIT) (No. RS-2022-NR068754), by the NVIDIA Academic Grant Program using NVIDIA RTX PRO 6000 GPUs and CUDA, and by the supercomputing resources provided by the Urban Big Data and AI Institute of the University of Seoul (\url{http://ubai.uos.ac.kr}).

\bibliography{custom}

\appendix

\clearpage

\section{Appendix}
\label{sec:appendix}

\subsection{Data Construction Details}
\label{app:data}

\textbf{Source Separation.}
The original HoloAssist recordings contain a single mono audio channel in which user speech, instructor speech, and background sounds are mixed together. Since all voices are pitch-shifted for anonymization, they cannot be used directly for training a natural-sounding spoken assistant. We therefore isolate only the background sounds using SAM-Audio (large-tv variant\footnote{\url{https://huggingface.co/facebook/sam-audio-large-tv}}), a prompt-based source separation model that takes an audio mixture and a reference audio prompt, extracts the sound matching the prompt, and returns both the extracted target and its residual. 

Since our goal is to preserve the background sounds while removing all human voices, we provide the speech segments as prompts so that SAM-Audio extracts and discards them, and we retain the residual as the background channel. For each speech-containing interval identified from the HoloAssist metadata timestamps, we supply the raw voice segment with a $\pm$0.2 second margin as the conditioning region to avoid sharp boundary artifacts in the separated output. To verify that no human voice remains in the resulting background, we apply Silero VAD to the separated background and discard all sessions where residual speech is still detected.

\textbf{Text Normalization.}
To synthesize user and assistant speech that is free of pitch shifting, preserves speaker anonymity, and is separated into distinct audio channels suitable for training, we first normalize the written-form dialog transcripts available in HoloAssist's metadata using a two-stage LLM pipeline with \textit{gpt-5.4-mini}. In Stage~A, a generator model receives the full conversation session and applies a text normalization covering spelling correction, punctuation standardization, canonical filler-word handling, and proper-noun formatting. The original HoloAssist conversations also contain task-irrelevant opening and closing remarks related to starting or stopping the recording (e.g., ``You can start recording now'' or ``We are done, you can stop''). These are identified using per-utterance metadata flags from the original annotations and removed, accounting for 9.0\% of all utterances. In Stage~B, an independent reviewer model (\textit{gpt-5.4-mini}) re-checks the Stage~A output for faithfulness and consistency.
\begin{table}[t]
\centering
\small
\begin{tabular}{lc}
\toprule
\textbf{Stage} & \textbf{Sessions} \\
\midrule
Original HoloAssist (train + val) & 1,673 \\
After source separation (VAD filter) & 1,532 \\
After TTS/WER filter + duration ($\geq$60s) & 1,419 \\
\midrule
\quad Final train split & 1,249 \\
\quad Final eval split & 170 \\
\bottomrule
\end{tabular}
\caption{Dataset attrition through the construction pipeline. Filters: (i) Silero VAD with \texttt{leak\_final=0} to remove source-separation residue, (ii) TTS/WER quality + minimum duration ($\geq$60s). The eval split consists of the surviving sessions from the original HoloAssist evaluation split and is not further filtered.}
\label{tab:attrition}
\vskip -10pt
\end{table}

\begin{figure*}[pt]
\centering
\includegraphics[width=\textwidth]{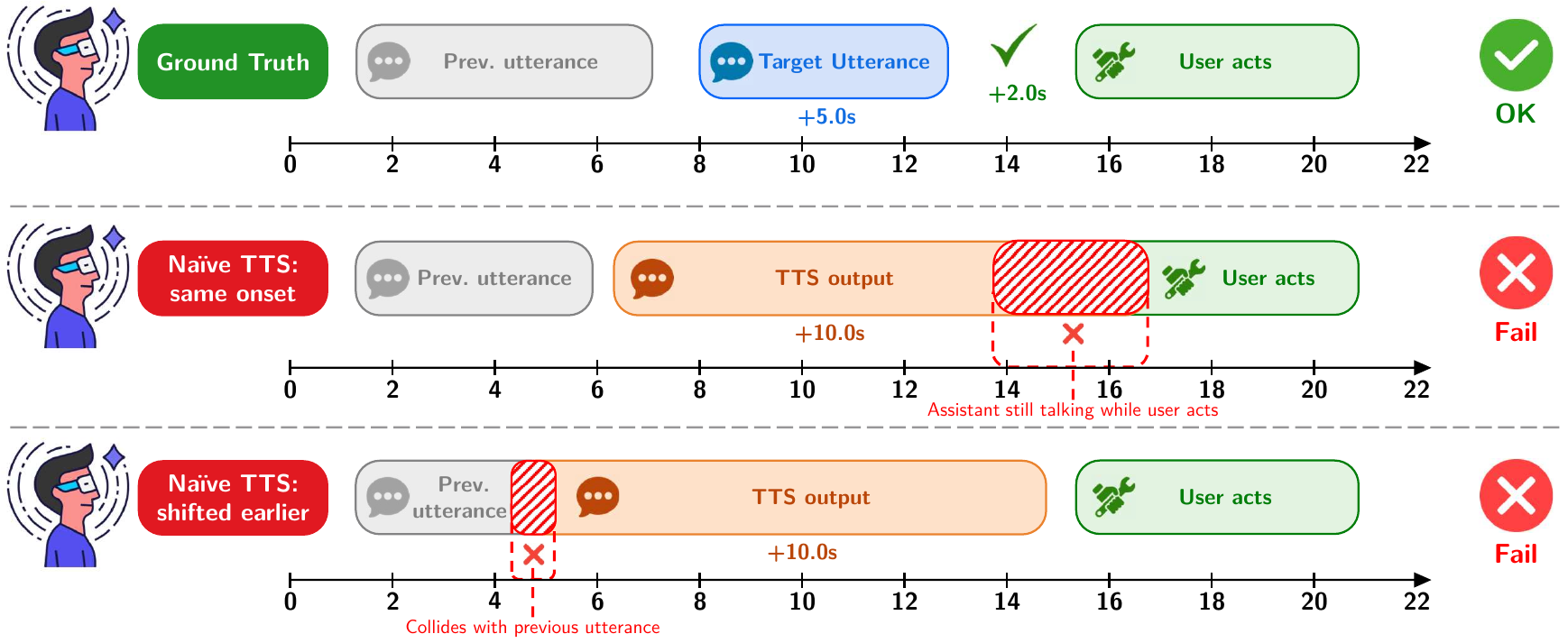}
\caption{Why naive TTS synthesis on video-text dialog data is insufficient. The ground truth instructor (top) times utterances to finish before the user's next action. Naive TTS (middle) generates longer speech that overlaps with the user's action. Shifting the onset earlier (bottom) avoids the action overlap but collides with the previous utterance and delivers guidance before the user has seen the relevant context. This motivates our choice to start from HoloAssist, where real instructor speech timing is already aligned with user actions, and to resynthesize speech with a strict duration constraint that preserves this alignment.}
\label{fig:tts_problem}
\vskip -10pt
\end{figure*}

\textbf{Speech Resynthesis.}
With the transcripts normalized for synthesis, we resynthesize clean speech using IndexTTS2~\cite{Zhou_Zhou_He_Zhou_Wang_Deng_Shu_2026}, a speaker-adaptive zero-shot TTS model. For each session, we assign one reference speaker to the user and one to the instructor, drawn from the English subset of Multilingual LibriSpeech~\cite{pratap20_interspeech} (2,498 training speakers, 340 disjoint evaluation speakers). Each utterance is synthesized 20 times, and candidates longer than the original utterance duration are discarded to prevent the synthesized speech from extending into the time window where the user performs the next physical action. From the remaining candidates, we transcribe each with Qwen3-ASR-1.7B~\cite{shi2026qwen3asrtechnicalreport} and select the one with the lowest word error rate (WER) against the normalized transcript. Sessions whose selected utterances all achieve WER = 0.0\% are automatically retained; all others are manually checked, and any session containing even one incorrectly pronounced utterance is discarded.

\textbf{Dataset Attrition.}
Table~\ref{tab:attrition} summarizes the number of sessions retained at each stage of the data construction pipeline. The largest reduction occurs at the source separation stage, where 141 sessions are discarded due to residual speech detected in the separated background.

\subsection{Baseline Configurations}
\label{app:baselines}

\textbf{Gemini Live API.} The Gemini Live API is Google's low-latency real-time API for streaming audio, video frames, and text, generating spoken responses. We use \texttt{gemini-live-2.5-flash-native-audio}\footnote{\url{https://docs.cloud.google.com/gemini-enterprise-agent-platform/models/gemini/2-5-flash-live-api}} through Vertex AI with JPEG video frames at 1 fps resized to 768$\times$768, context-window compression and proactive audio enabled, and a 15-second tail wait after all video frames are sent to allow the model to finish generating.

\textbf{OpenAI Realtime API.} The OpenAI Realtime API enables low-latency speech-to-speech interaction with GPT-class models over WebSocket. We evaluate two snapshots, \texttt{gpt-realtime}\footnote{\url{https://developers.openai.com/api/docs/models/gpt-realtime}} and \texttt{gpt-realtime-2}\footnote{\url{https://developers.openai.com/api/docs/models/gpt-realtime-2}}, with the \texttt{alloy} voice, server-side VAD for turn detection, and video frames at 1 fps (768 px). Input audio is 16-bit PCM at 24 kHz.

\begin{table*}[pt]
\centering
\small
\begin{tabular}{llcc}
\toprule
& \textbf{Hyperparameter} & \textbf{SFT} & \textbf{DPO} \\
\midrule
\multirow{3}{*}{\textbf{\textit{LoRA}}}
& Rank ($r$) / Alpha ($\alpha$) / Dropout & \multicolumn{2}{c}{32 / 64 / 0.05} \\
& Target modules & All linear (LLM + TTS) & All linear (LLM only) \\
& TTS decoder & Trained & Frozen \\
\midrule
\multirow{2}{*}{\textbf{\textit{Loss}}}
& Objective & $\mathcal{L}_{\text{LM}} + \lambda \mathcal{L}_{\text{S3}}$ & Length-norm. DPO \\
& Key parameter & $\lambda = 0.1$, detach = True & $\beta = 0.1$ \\
\midrule
\multirow{5}{*}{\textbf{\textit{Optimization}}}
& Learning rate & 2e-4 & 5e-6 \\
& Schedule & \multicolumn{2}{c}{Cosine with 5\% warmup} \\
& Weight decay & 0.01 & 0.0 \\
& Max gradient norm & \multicolumn{2}{c}{1.0} \\
& Effective batch size & 32 & 16 \\
\midrule
\multirow{2}{*}{\textbf{\textit{Training}}}
& Epochs / steps & 5 / 2,140 & 1 / 334 \\
& Data per epoch & 13,676 windows & 5,335 pairs \\
\midrule
\multirow{1}{*}{\textbf{\textit{Infra.}}}
& Precision & \multicolumn{2}{c}{bf16 + FlashAttention-2~\cite{dao2024flashattention}} \\
\bottomrule
\end{tabular}
\caption{Training hyperparameters for SFT and DPO.}
\label{tab:hparams}
\vskip -10pt
\end{table*}

Both the Gemini Live API and OpenAI Realtime API rely on voice activity detection to trigger responses, meaning they respond primarily when the user speaks. Since users are largely silent during physical tasks in our setting, these models rarely intervene: the average number of utterances per session is 0.25 for Gemini Live, 0.79 for \texttt{gpt-realtime-2}, and 0.94 for \texttt{gpt-realtime}. We also remove background noise from the input audio for these models to prevent false VAD triggering, providing a more favorable evaluation condition, though the results suggest that these models are not designed for proactive assistance in this setting.

\textbf{ProAssist.} ProAssist~\cite{zhang-etal-2025-proactive} is an 8B open-source video-text model built on VideoLLM-Online~\cite{Chen_2024_CVPR} that learns proactive dialog generation from streaming egocentric video. It handles text rather than speech and does not accept audio input. We use the default settings from the original paper and official implementation, with native 2 fps video at 384$\times$384 resolution. In the cascade variant, the ground-truth user transcripts are replaced by Qwen3-ASR-1.7B transcripts of the resynthesized user speech, and ProAssist's text output is synthesized with IndexTTS2.

\textbf{MiniCPM-o 4.5.} MiniCPM-o 4.5~\cite{cui2026minicpmo45realtimefullduplex} is a 9B open-source omni-modal model that supports real-time full-duplex interaction across vision, speech, and text. It serves as our backbone before fine-tuning. We evaluate it in a zero-shot setting with a system prompt that provides the task name and step-by-step procedural instructions, using temperature = 0.7, top-$p$ = 0.8, and top-$k$ = 100. Unlike the closed-source APIs that rarely speak, MiniCPM-o 4.5 generates an average of 4.85 utterances per session. However, an analysis of all 825 utterances across the 170 benchmark sessions reveals that only 30.1\% constitute proactive guidance (Table~\ref{tab:utttype}). The remaining utterances are third-person scene narration (21.0\%, e.g., ``The person is unscrewing the leg''), which is inappropriate for a first-person assistant, or acknowledgements, greetings, and hallucinated text (49.0\%, e.g., ``Is Jesus Christ'', ``Oh'', or garbled tokens).

\subsection{Training Hyperparameters}
\label{app:hparams}

We provide the full hyperparameter settings for SFT and DPO training in Table~\ref{tab:hparams}. All results reported in this paper are from single training runs. Training and evaluation are performed using three GPU types: NVIDIA A6000, RTX PRO 6000, and B200.

\subsection{Evaluation Details}
\label{app:eval}

\textbf{LLM Judge.} We prompt \textit{gpt-5.4-mini}\footnote{\url{https://developers.openai.com/api/docs/models/gpt-5.4-mini}} (reasoning effort = medium) with the full session context, including all reference and predicted utterances with their timestamps. Our judge prompt and scoring rubric are largely adapted from ProAssist~\cite{zhang-etal-2025-proactive}. The judge rates each session on four axes on a 1-to-5 Likert scale: Correctness, Promptness, Efficiency, and Overall. The full system prompt is shown in Figure~\ref{fig:judge_prompt}.

\begin{figure*}[p]
\begin{tcolorbox}[colback=gray!5, colframe=gray!60, title={\textbf{LLM Judge System Prompt} (\textit{gpt-5.4-mini}, reasoning effort = medium)}, fonttitle=\small, fontupper=\footnotesize, left=4pt, right=4pt, top=2pt, bottom=2pt]
\texttt{You are an expert in evaluating the quality of user-assistant dialogues. Your task is to evaluate dialog responses generated by an assistant model that helps users with their tasks. You should evaluate the dialogs by comparing them to reference gold-standard dialogues from professional assistants.}

\vspace{4pt}
\texttt{Requirement:}\\
\texttt{1. Read dialogues carefully and compare them line by line. Keep your analysis concise and to the point.}\\
\texttt{2. Evaluate the following aspects:}\\
\texttt{\quad - Correctness: does each generated instruction/feedback make sense (correct or relevant) or not, based on the context and the gold-standard reference?}\\
\texttt{\quad - Promptness: does the assistant provide guidance at the right time, or does it talk too early or too late?}\\
\texttt{\quad - Efficiency: does the assistant provide the necessary information in a concise and efficient manner, without too much repetition or redundancy information?}\\
\texttt{\quad - Overall: the overall helpfulness and quality of the assistant's responses.}\\
\texttt{3. For each aspect, give a score from 1 to 5 based on the following criteria:}\\
\texttt{\quad - 1=very poor: most of utterances are incorrect, irrelevant, mistimed, inefficient etc}\\
\texttt{\quad - 2=poor: bad utterances that are incorrect, irrelevant, mistimed are more than good ones}\\
\texttt{\quad - 3=average: the number of good and bad utterances are roughly the same}\\
\texttt{\quad - 4=good: more good utterances than bad ones}\\
\texttt{\quad - 5=excellent: most of utterances are correct, relevant, timely, efficient etc}

\end{tcolorbox}
\caption{System prompt used for LLM-as-a-judge evaluation. The model is forced to emit a strict JSON schema with five keys: \texttt{analysis}, \texttt{correctness}, \texttt{promptness}, \texttt{efficiency}, \texttt{overall} (latter four integers 1-5).}
\label{fig:judge_prompt}
\vskip -10pt
\end{figure*}

\begin{figure*}[p]
\begin{tcolorbox}[colback=blue!3, colframe=blue!40, title={\textbf{Action-Linked Classification Prompt} (\textit{gpt-5.5}, reasoning effort = high, temperature $=1.0$)}, fonttitle=\small, fontupper=\footnotesize, left=4pt, right=4pt, top=2pt, bottom=2pt]
\texttt{You are classifying utterances from an AI assistant guiding a user through a physical procedure (assembly, cooking, etc.) in an egocentric task-guidance setting.}

\vspace{4pt}
\texttt{For each utterance you receive:}\\
\texttt{\quad - the utterance text}\\
\texttt{\quad - a $\pm$15s recipe context window (the user's nearby motor actions and other assistant utterances, each with a timestamp).}

\vspace{4pt}
\texttt{Decide whether the utterance is a DIRECTIVE: an instruction that tells the user to start a physical motor action they have not yet started.}

\vspace{4pt}
\texttt{INCLUDE as directive:}\\
\texttt{\quad - Imperative: ``Pull the lever'', ``Take out the battery''}\\
\texttt{\quad - Permission-framed instruction: ``Now you can change the battery''}\\
\texttt{\quad - Conditional instruction: ``If the screw is loose, tighten it''}\\
\texttt{\quad - Action reminders: ``Don't forget the cap''}

\vspace{4pt}
\texttt{EXCLUDE (not directive):}\\
\texttt{\quad - Acknowledgement: ``Okay'', ``Yes'', ``Great'', ``Thank you'', ``Mm-hm''}\\
\texttt{\quad - Confirmation: ``That's right'', ``Correct''}\\
\texttt{\quad - Commentary / observation: ``It's fine'', ``It looks loose''}\\
\texttt{\quad - Pure deictic without action verb: ``This one'', ``Yes, here''}\\
\texttt{\quad - Questions: ``Did you find it?''}\\
\texttt{\quad - Greetings / closing remarks: ``We are done'', ``Nice work''}

\vspace{4pt}
\texttt{Use "ambiguous" only when genuinely undecidable from text + context.}

\vspace{4pt}
\texttt{Respond as JSON exactly matching the schema. Keep rationale $\leq$ 30 words.}

\end{tcolorbox}
\caption{Prompt used for action-linked utterance classification with majority voting (strict-majority over 5 independent calls at temperature $1.0$; ties resolved by the most frequent label). Response is enforced as strict JSON \texttt{\{is\_directive $\in$ \{yes, no, ambiguous\}, rationale\}}.}
\label{fig:classifier_prompt}
\vskip -10pt
\end{figure*}

\begin{figure*}[tp]
\begin{tcolorbox}[colback=yellow!8, colframe=olive!60, title={\textbf{Utterance Type Classification Prompt} (\textit{gpt-5.6-luna}, 5-way majority voting)}, fonttitle=\small, fontupper=\footnotesize, left=4pt, right=4pt, top=2pt, bottom=2pt]
\texttt{You label single utterances spoken during a procedural physical task (assembling furniture, operating a camera, making coffee...). The speaker guides a person who is performing the task and who is wearing a head-mounted camera. The speaker may be a human instructor or an AI assistant; label the utterance itself and do not speculate about who produced it.}

\vspace{4pt}
\texttt{Assign exactly one label.}

\vspace{4pt}
\texttt{GUIDANCE: the utterance tells the person what to do next, how to do it, what to avoid, or what to fix. The person could act on it. This includes instructions, suggestions, warnings, corrections, and reminders.}\\
\texttt{\quad ``Now attach the leg to the corner bracket.''}\\
\texttt{\quad ``Careful, that screw goes in the other way.''}\\
\texttt{\quad ``You will need the small hex key for this one.''}\\
\texttt{\quad ``Try turning it counterclockwise instead.''}

\vspace{4pt}
\texttt{NARRATION: the utterance describes the scene, an object, or what the person is doing, without asking for any action. Typically third person or purely observational.}\\
\texttt{\quad ``The person is holding a screwdriver.''}\\
\texttt{\quad ``There is a box of screws on the table.''}\\
\texttt{\quad ``He is removing the lens cover.''}

\vspace{4pt}
\texttt{OTHER: anything else. This includes acknowledgments and praise with no action attached (``Okay'', ``Great job'', ``Mhm''), confirmations and answers (``Yes'', ``That one''), greetings and closings, meta-talk about the conversation or recording, questions that request information rather than give it, filler, and text that is unintelligible or in the wrong language.}

\vspace{4pt}
\texttt{Boundary rules.}\\
\texttt{1. Praise or acknowledgment alone is OTHER. It becomes GUIDANCE only when it is attached to an instruction (``Good, now tighten it'').}\\
\texttt{2. A description that implies an obvious next action is still NARRATION unless the action is actually stated.}\\
\texttt{3. Judge only the text given. Do not infer intent from surrounding context you cannot see.}

\vspace{4pt}
\texttt{Return strict JSON: \{``label'': ``GUIDANCE'' | ``NARRATION'' | ``OTHER'', ``reason'': ``$\leq$ 15 words''\}}
\end{tcolorbox}
\caption{Prompt used for utterance type classification. The label is the strict majority over 5 independent calls.}
\label{fig:utttype_prompt}
\vskip -10pt
\end{figure*}

\textbf{Action-Linked Classification.} To construct the action-linked evaluation subset for end-vs-action timing, we classify each reference utterance as either a procedural directive (e.g., ``change the battery'') or a non-directive remark (e.g., acknowledgement, encouragement) using \textit{gpt-5.5}\footnote{\url{https://developers.openai.com/api/docs/models/gpt-5.5}} (reasoning effort = high) with 5-way self-consistency majority voting at temperature = 1.0. Procedural directives are linked to the user's nearest subsequent physical action from HoloAssist's action annotations within a temporal window of $T$ = 5 seconds. This retains 942 out of 2,067 total reference utterances across 170 benchmark sessions. The classification prompt is shown in Figure~\ref{fig:classifier_prompt}.

\textbf{Utterance Type Classification.} To characterize what kind of speech each system generates (Table~\ref{tab:utttype}), we classify every delivered utterance, and every reference instructor utterance, as proactive guidance, third-person scene narration, or other. We use \textit{gpt-5.6-luna}\footnote{\url{https://developers.openai.com/api/docs/models/gpt-5.6-luna}} with the prompt in Figure~\ref{fig:utttype_prompt}, taking the strict majority over 5 independent calls. The same prompt is applied to model output and to reference utterances so that the two are directly comparable. Across the four sources, 91.6\% to 96.3\% of utterances received a unanimous label.

\textbf{Content Similarity.} We compute cosine similarity between reference and predicted utterances using Qwen3-Embedding-0.6B~\cite{zhang2025qwen3embeddingadvancingtext}. For each matched reference-prediction pair identified by the onset timing metric, we encode both utterances and compute their cosine similarity as a measure of semantic alignment.

\textbf{ASR Transcription.} For the \texttt{text-asr} evaluation channel, we transcribe the generated speech using Qwen3-ASR-1.7B~\cite{shi2026qwen3asrtechnicalreport}. To determine utterance boundaries within continuous speech output, we apply Qwen3-ForcedAligner-0.6B~\cite{shi2026qwen3asrtechnicalreport} for word-level timestamp alignment, followed by Silero VAD (threshold = 0.5)~\cite{SileroVAD} for utterance segmentation. All artifacts and their licenses are listed in Table~\ref{tab:artifacts} (Appendix~\ref{app:licenses}).

\textbf{End-of-Utterance Decomposition.} On the action-linked subset, every matched prediction has a user action attached to it through the reference utterance it matched. We measure three things against the start of that action: when the predicted utterance begins (start-vs-action), when it ends (end-vs-action), and how long it lasts. Both the onset and the end are read from the generated audio: we anchor the utterance at the chunk in which the model starts speaking and locate its first and last word with Qwen3-ForcedAligner-0.6B. Its duration is therefore the span over which the model is audible, which is shorter than the synthesized audio file since that also contains leading and trailing silence. The three quantities satisfy end-vs-action = start-vs-action - duration for every pair, and hence for their means, though not for their medians. The per-model decomposition is reported in Appendix~\ref{app:enddist}.

\subsection{Human Evaluation}
\label{app:human_eval}

The full human evaluation results are described in Section~\ref{sec:human_eval}. Each clip is centered on a ground truth instructor utterance and the user action it prompts, with temporal margins on both sides for context. We remove the human instructor's speech and overlay the spoken responses of two anonymous assistants, keeping the original user speech and background audio, so that the two versions of a clip differ only in the assistant.

\begin{figure*}[p]
\centering
\includegraphics[height=0.88\textheight]{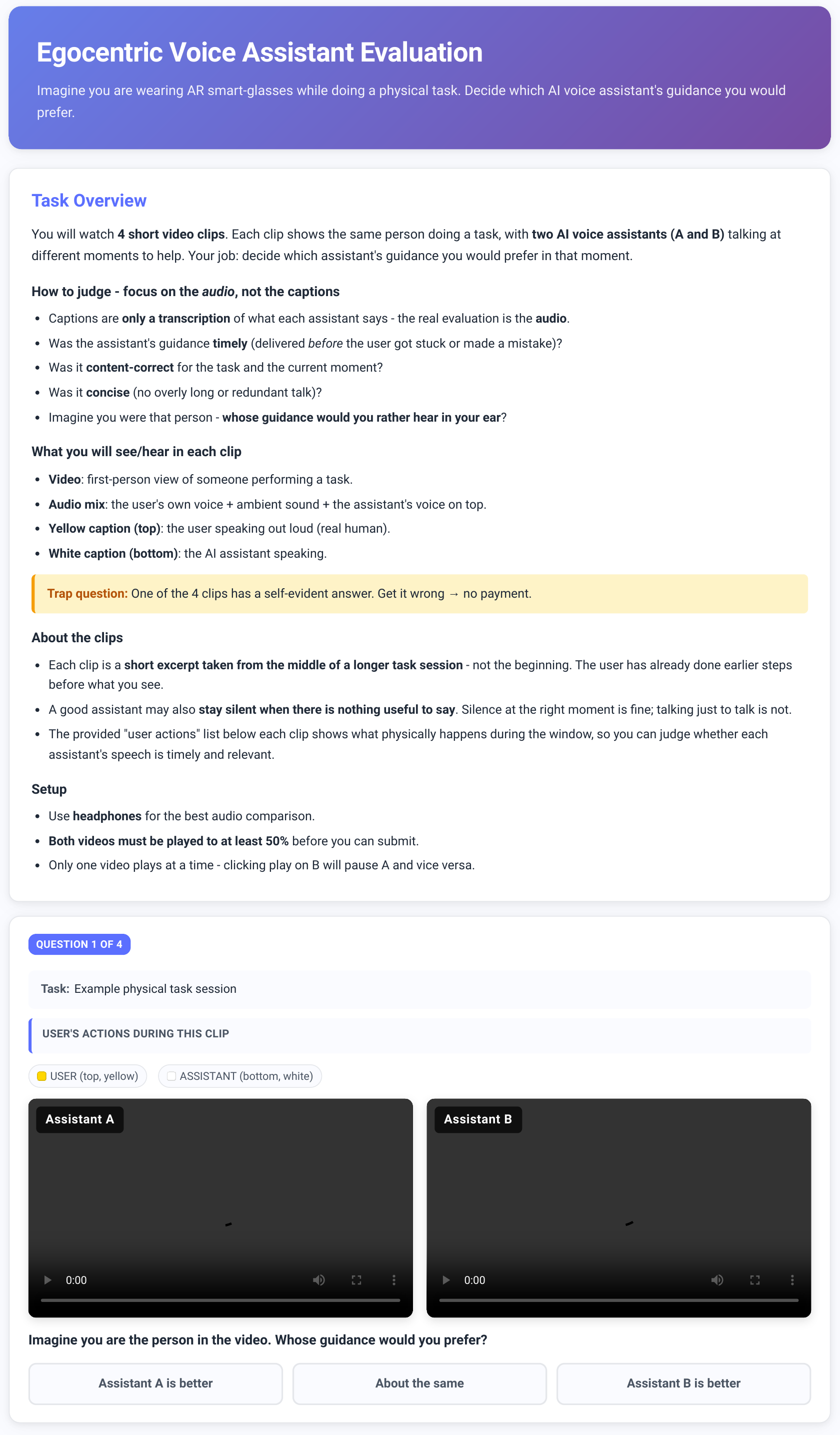}
\caption{Human evaluation interface on Amazon Mechanical Turk. Evaluators watch two versions of the same egocentric clip with different assistant speech overlays and synchronized captions, then select which assistant (A or B) provides more helpful and timely guidance, or indicate a tie.}
\label{fig:human_eval_interface}
\vskip -10pt
\end{figure*}

We recruit 100 workers per comparison on Amazon Mechanical Turk. Each HIT contains three real A/B comparisons and one attention test (ground truth vs silence); workers who select the wrong answer on the test are excluded from the final scoring, leaving 252--267 valid judgments per comparison. A/B sides are randomized per question, and workers must achieve at least 50\% cumulative playback on both clips before submitting. The total cost across all three comparisons is approximately \$200. The evaluation template is shown in Figure~\ref{fig:human_eval_interface}.

\begin{table*}[t]
\centering
\small
\begin{tabular}{l ccc c cccc c}
\toprule
\multirow{2}{*}{\textbf{Model}} & \multicolumn{3}{c}{\textbf{Onset Timing}} & \textbf{Content} & \multicolumn{4}{c}{\textbf{LLM Judge (1--5)}} & \textbf{End Timing} \\
\cmidrule(lr){2-4} \cmidrule(lr){5-5} \cmidrule(lr){6-9} \cmidrule(lr){10-10}
& \textbf{Prec.} & \textbf{Rec.} & \textbf{F1} & \textbf{Cosine} & \textbf{Correct.} & \textbf{Prompt.} & \textbf{Effic.} & \textbf{Overall} & \textbf{end-vs-act (s)} \\
\midrule
Gemini Live & .364 & .019 & .022 & .453 & 2.24 & 2.39 & 2.82 & 2.06 & $-$3.65 / $-$3.62 \\
\texttt{gpt-realtime} & .285 & .024 & .030 & .499 & 2.24 & 2.70 & 2.54 & 2.20 & $-$7.40 / $-$7.46 \\
\texttt{gpt-realtime-2} & .256 & .019 & .024 & .457 & 2.32 & 2.30 & 2.22 & 2.14 & $-$16.0 / $-$16.8 \\
\midrule
ProAssist$^\dagger$ & .480 & .158 & .195 & .594 & 2.34 & 2.63 & 2.99 & 2.06 & -- \\
\quad + ASR/TTS cascade & .479 & .158 & .195 & .585 & 2.35 & 2.63 & 2.97 & 2.07 & $-$0.44 / $-$1.56 \\
MiniCPM-o 4.5 & .296 & .111 & .122 & .538 & 1.76 & 1.73 & 1.99 & 1.58 & $-$4.81 / $-$1.18 \\
\midrule
EgoVoice-SFT & .488 & .241 & .284 & .707 & 2.59 & 2.74 & 2.77 & 2.37 & +1.42 / +1.10 \\
EgoVoice-DPO & .521 & .302 & .346 & .674 & 2.61 & 2.68 & 2.71 & 2.39 & +1.25 / +1.04 \\
\bottomrule
\end{tabular}
\caption{Results on 170 benchmark sessions (\texttt{text-llm}). $\dagger$~indicates a text-only baseline. On this channel the cascade row is ProAssist reading ASR transcripts of the user's speech instead of ground-truth transcripts, so the two rows differ only in that input. Notation otherwise follows Table~\ref{tab:main}.}
\label{tab:textllm}
\end{table*}

\begin{table*}[t]
\centering
\small
\begin{tabular}{lrccc}
\toprule
\textbf{Model} & \textbf{n} & \textbf{Duration} & \textbf{Start-vs-action (s)} & \textbf{End-vs-action (s)} \\
\midrule
Gemini Live & 3 & 6.64 / 6.48 & +3.44 / +3.04 & $-$3.20 / $-$2.26 \\
\texttt{gpt-realtime} & 30 & 11.73 / 10.84 & +4.34 / +4.05 & $-$7.40 / $-$7.46 \\
\texttt{gpt-realtime-2} & 22 & 20.39 / 20.20 & +4.33 / +4.29 & $-$16.1 / $-$16.8 \\
\midrule
ProAssist + ASR/TTS cascade & 92 & 6.38 / 6.20 & +5.94 / +4.82 & $-$0.44 / $-$1.56 \\
MiniCPM-o 4.5 & 99 & 9.87 / 6.00 & +5.18 / +4.70 & $-$4.69 / $-$0.40 \\
\midrule
EgoVoice-SFT & 255 & 2.99 / 2.32 & +4.55 / +4.38 & +1.57 / +1.52 \\
EgoVoice-DPO & 287 & 2.93 / 2.56 & +4.68 / +4.30 & +1.75 / +1.62 \\
\midrule
Human instructor & 942 & 3.05 / 2.60 & +4.74 / +4.37 & +1.69 / +1.38 \\
\bottomrule
\end{tabular}
\caption{Decomposition of end-vs-action on the action-linked subset ($T{=}5$), in seconds, as mean / median. $n$ is the number of matched interventions, which differs across models.}
\label{tab:decomp}
\end{table*}

\begin{figure*}[t]
\centering
\includegraphics[width=0.95\textwidth]{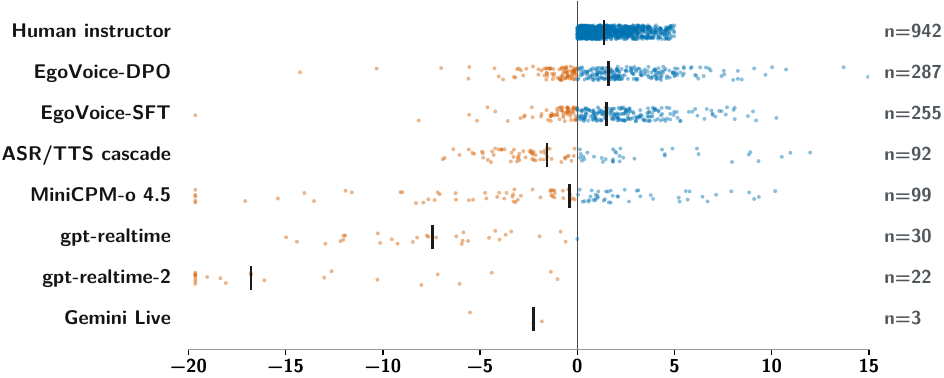}
\caption{End-vs-action for every matched action-linked intervention ($T{=}5$, \texttt{text-asr}), in seconds. Each dot is one intervention; blue means the model finished speaking before the user began the linked action, orange means it was still speaking. The vertical tick marks the median and $n$ is the number of matched interventions. Values outside the plotted range are clipped to it.}
\label{fig:enddist}
\end{figure*}

\subsection{Additional Results}
\label{app:additional}

\subsubsection{Results on the \texttt{text-llm} Channel}
\label{app:textllm}

Table~\ref{tab:textllm} repeats the evaluation of Table~\ref{tab:main} on \texttt{text-llm}, the text a model generates internally, rather than \texttt{text-asr}, the transcription of the speech it actually generates. Most content and judge metrics are slightly higher on \texttt{text-llm} for models that generate speech, as the text is not subject to TTS or ASR noise. End-vs-action also shifts between the two channels. The end of an utterance is measured from the generated audio in both cases, so it does not change. The onset does: on \texttt{text-asr} it is where the speech is first heard, and on \texttt{text-llm} it is the chunk where the model emits the text. Because predictions are matched to references by onset, the two channels score slightly different sets of interventions. The cascade row also serves as an ASR control here: it is ProAssist reading Qwen3-ASR-1.7B transcripts of the user's speech instead of ground-truth ones, and onset F1 is unchanged at 0.195 while cosine drops by only 0.009. The relative ranking is largely consistent between the two channels. Audio samples of our dataset, qualitative comparisons between EgoVoice and baselines, and additional examples are available on the project page.\footnote{\url{https://egocentricvoice.github.io/}}

\subsubsection{End-of-Utterance Decomposition}
\label{app:enddist}

Table~\ref{tab:decomp} separates end-vs-action into the two quantities that produce it. Every system starts speaking at a broadly similar point relative to the user's action, between +3.4 and +5.9 seconds. What differs is how long each one talks. EgoVoice-SFT and EgoVoice-DPO speak for 2.99 and 2.93 seconds on average, close to the instructor's 3.05, while the cascade takes 6.38 seconds, MiniCPM-o 4.5 takes 9.87, and \texttt{gpt-realtime-2} takes 20.39. Since end-vs-action is start-vs-action minus this duration, utterance length alone accounts for the sign of the result. Deciding when to speak is not the difficult part; finishing in time is.

Figure~\ref{fig:enddist} shows the individual interventions behind these numbers. The instructor almost never runs late, and the two EgoVoice variants have the same shape with a heavier left tail. The baselines differ in kind rather than degree: the cascade and MiniCPM-o 4.5 sit astride zero, and the three closed-source APIs fall almost entirely below it. The figure also shows why we report medians alongside means. MiniCPM-o 4.5 has a mean of $-4.69$ but a median of only $-0.40$, because a handful of very late endings pull the average down.

\subsubsection{Utterance Statistics}
\label{app:uttstats}

Table~\ref{tab:uttstats} reports how often each system speaks and how much it says. The two EgoVoice variants match the instructor's turn structure closely, at 8.1 and 7.8 words per utterance against 8.3, with a similar share of brief acknowledgements. They still speak about half as often as the instructor. Every baseline takes far longer turns, from 16.6 words for ProAssist to 51.6 for \texttt{gpt-realtime-2}, and almost never produces a short acknowledgement. This is the turn structure that the Efficiency axis of the LLM judge rewards, and it is also what pushes the baselines past the user's next action in Table~\ref{tab:decomp}. Counts here are on \texttt{text-llm}, so they are lower than the \texttt{text-asr} counts quoted in Appendix~\ref{app:baselines} for MiniCPM-o 4.5.
  
\subsubsection{Significance Tests}
\label{app:significance}

Table~\ref{tab:significance} collects the tests reported in Section~\ref{sec:results}. Onset F1 separates every pair of systems we compare. The LLM judge separates both EgoVoice variants from the zero-shot backbone, but neither from ProAssist nor from each other. We do not test cosine similarity. Unlike the other two metrics, it is defined per matched pair, and each model is scored on a different set of pairs, so there is no common set on which to run a paired test.

\begin{table}[t]
\centering
\small
\setlength{\tabcolsep}{4.5pt}
\begin{tabular}{lccc}
\toprule
\textbf{Model} & \textbf{Utt.} & \textbf{Words} & \textbf{$\leq$3 words} \\
\midrule
Gemini Live & 0.3 & 24.2 & 2.3\% \\
\texttt{gpt-realtime} & 0.9 & 37.7 & 0.0\% \\
\texttt{gpt-realtime-2} & 0.8 & 51.6 & 0.0\% \\
\midrule
ProAssist$^\dagger$ & 2.2 & 16.6 & 0.8\% \\
\quad + ASR/TTS cascade & 2.1 & 16.7 & 1.9\% \\
MiniCPM-o 4.5 & 3.4 & 38.2 & 8.2\% \\
\midrule
EgoVoice-SFT & 4.6 & 8.1 & 20.2\% \\
EgoVoice-DPO & 5.6 & 7.8 & 19.3\% \\
\midrule
Human instructor & 11.3 & 8.3 & 17.9\% \\
\bottomrule
\end{tabular}
\caption{Utterance statistics on the \texttt{text-llm} channel: utterances per session, mean words per utterance, and the share of utterances of three words or fewer.}
\label{tab:uttstats}
\end{table}

\begin{table}[t]
\centering
\small
\setlength{\tabcolsep}{4.5pt}
\begin{tabular}{lrrl}
\toprule
\textbf{Comparison} & $\boldsymbol{n}$& \textbf{Diff} & \textbf{\textit{p}} \\
\midrule
\multicolumn{4}{l}{\textbf{\textit{Onset F1}}} \\
DPO vs SFT & 170 & $+$0.050 & 0.0007 \\
DPO vs MiniCPM & 170 & $+$0.211 & 6.0e$-$19 \\
SFT vs MiniCPM & 170 & $+$0.161 & 1.2e$-$13 \\
DPO vs ProAssist & 170 & $+$0.129 & 1.1e$-$08 \\
SFT vs ProAssist & 170 & $+$0.079 & 7.9e$-$05 \\
\midrule
\multicolumn{4}{l}{\textbf{\textit{LLM Judge (Overall)}}} \\
DPO vs SFT & 152 & $-$0.072 & 0.42 \\
DPO vs MiniCPM & 156 & $+$0.776 & 1.1e$-$13 \\
SFT vs MiniCPM & 149 & $+$0.832 & 3.6e$-$13 \\
DPO vs ProAssist & 160 & $+$0.094 & 0.38 \\
SFT vs ProAssist & 153 & $+$0.157 & 0.12 \\
\bottomrule
\end{tabular}
\caption{Paired Wilcoxon signed-rank tests over sessions. Onset F1 is defined for all 170 sessions, while judge scores exist only for sessions in which both models speak, so $n$ varies. Comparisons against ProAssist use \texttt{text-llm} on both sides.}
\label{tab:significance}
\end{table}

\begin{figure*}[pt]
\centering
\includegraphics[width=\textwidth]{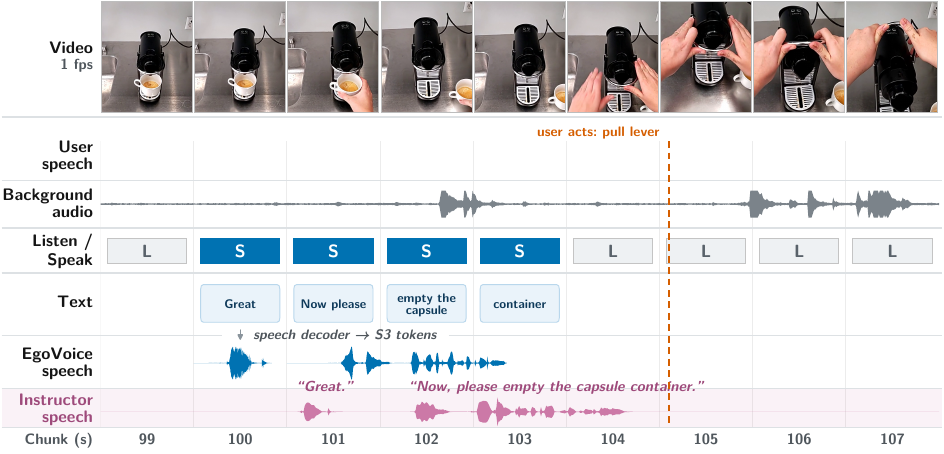}\\[50pt]
\includegraphics[width=\textwidth]{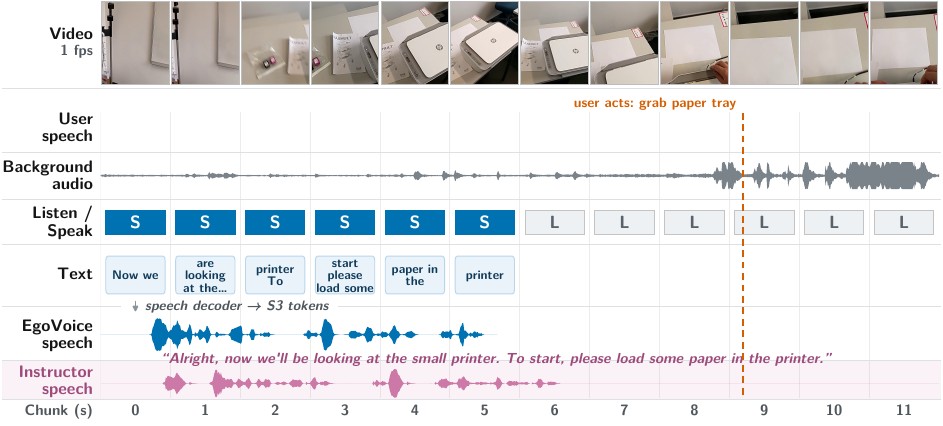}
\caption{Two interventions that succeed. Each panel follows Figure~\ref{fig:stream}, with the instructor's own speech added at the bottom (shaded) and the start of the user's linked action marked. EgoVoice-DPO speaks at the same moment as the instructor, says the same thing, and finishes before the user acts. The system prompt lists the task steps in third person, so the model has to choose which step is due, phrase it as an instruction, and time it.}
\label{fig:case_success}
\end{figure*}

\begin{figure*}[pt]
\centering
\includegraphics[width=\textwidth]{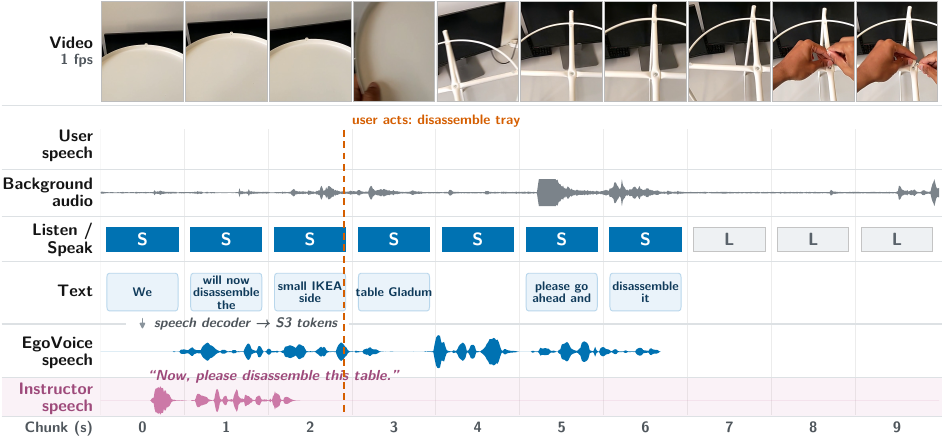}\\[50pt]
\includegraphics[width=\textwidth]{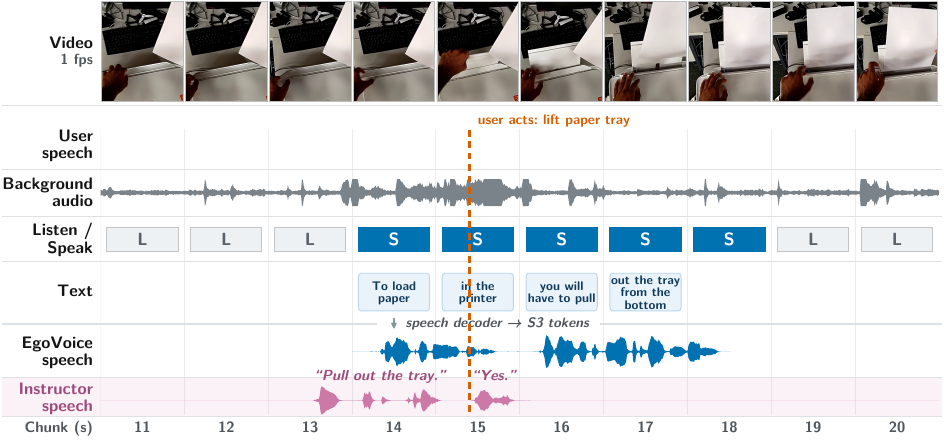}
\caption{Two interventions that end too late. The onset and the content are right, but the model's speech runs well past the instructor's and is still playing when the user begins the action.}
\label{fig:case_late}
\end{figure*}

\begin{figure*}[pt]
\centering
\includegraphics[width=\textwidth]{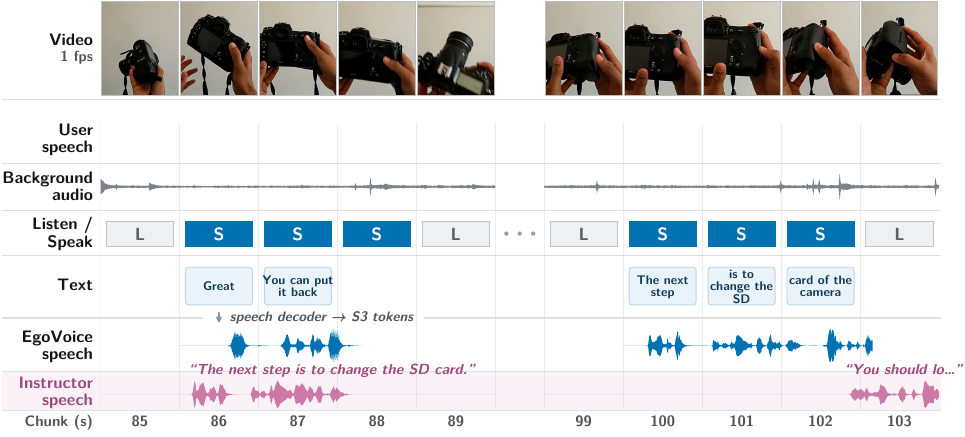}\\[50pt]
\includegraphics[width=\textwidth]{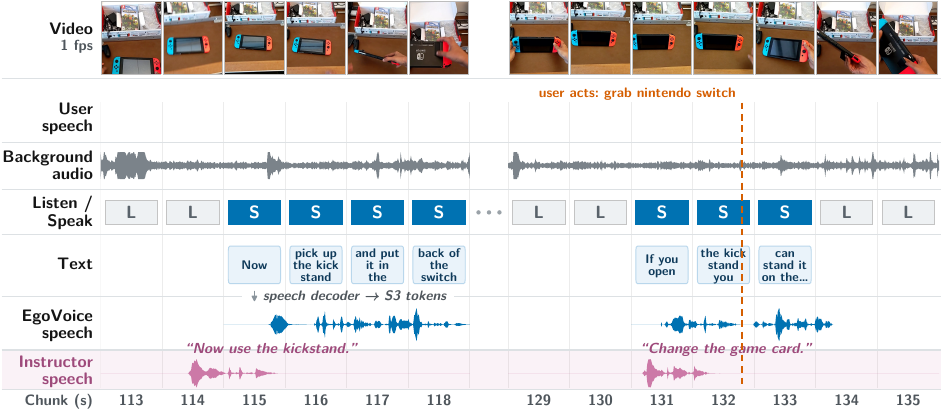}
\caption{Two interventions that guide the wrong step. The model delivers an instruction that belongs elsewhere in the task: in the first, one the instructor gave 14 seconds earlier; in the second, one the model itself gave 17 seconds earlier. In both cases the instructor has already moved on.}
\label{fig:case_step}
\end{figure*}

\begin{table*}[t]
\centering
\small
\begin{tabular}{lm{2.0cm}m{7.0cm}}
\toprule
\textbf{Artifact} & \textbf{License} & \textbf{URL} \\
\midrule
\multicolumn{3}{l}{\textbf{\textit{Datasets}}} \\
HoloAssist & CDLAv2 & \url{https://holoassist.github.io} \\
Multilingual LibriSpeech & CC BY 4.0 & \url{https://www.openslr.org/94} \\
\midrule
\multicolumn{3}{l}{\textbf{\textit{Backbone \& Training}}} \\
MiniCPM-o 4.5 & apache-2.0 & \url{https://huggingface.co/openbmb/MiniCPM-o-4_5} \\
\midrule
\multicolumn{3}{l}{\textbf{\textit{Data Construction}}} \\
IndexTTS2 & Bilibili Model License & \url{https://huggingface.co/IndexTeam/IndexTTS-2} \\
SAM-Audio (large-tv) & SAM License & \url{https://huggingface.co/facebook/sam-audio-large-tv} \\
Silero VAD & MIT & \url{https://github.com/snakers4/silero-vad} \\
\midrule
\multicolumn{3}{l}{\textbf{\textit{Evaluation}}} \\
Qwen3-ASR-1.7B & apache-2.0 & \url{https://huggingface.co/Qwen/Qwen3-ASR-1.7B} \\
Qwen3-ForcedAligner-0.6B & apache-2.0 & \url{https://huggingface.co/Qwen/Qwen3-ForcedAligner-0.6B} \\
Qwen3-Embedding-0.6B & apache-2.0 & \url{https://huggingface.co/Qwen/Qwen3-Embedding-0.6B} \\
\textit{gpt-5.4-mini} (judge) & Proprietary & \url{https://developers.openai.com/api/docs/models/gpt-5.4-mini} \\
\textit{gpt-5.5} (action-linked classifier) & Proprietary & \url{https://developers.openai.com/api/docs/models/gpt-5.5} \\
\textit{gpt-5.6-luna} (utterance type classifier) & Proprietary & \url{https://developers.openai.com/api/docs/models/gpt-5.6-luna} \\
\midrule
\multicolumn{3}{l}{\textbf{\textit{Baselines}}} \\
Gemini Live API & Proprietary & \url{https://docs.cloud.google.com/vertex-ai/generative-ai/docs/models/gemini/2-5-flash-live-api} \\
\texttt{gpt-realtime} & Proprietary & \url{https://developers.openai.com/api/docs/models/gpt-realtime} \\
\texttt{gpt-realtime-2} & Proprietary & \url{https://developers.openai.com/api/docs/models/gpt-realtime-2} \\
ProAssist & Not specified & \url{https://huggingface.co/594zyc/ProAssist-Model-L4096-I10} \\
\bottomrule
\end{tabular}
\caption{All artifacts used in this work with their licenses and sources.}
\label{tab:artifacts}
\vskip -10pt
\end{table*}

\subsubsection{Qualitative Examples and Failure Modes}
\label{app:qualitative}

Figures~\ref{fig:case_success} to \ref{fig:case_step} show six EgoVoice-DPO interventions in the format of Figure~\ref{fig:stream}, with two additions: the instructor's own speech is drawn at the bottom of each panel, so the two waveforms can be compared directly, and the start of the user's linked action is marked. Figure~\ref{fig:case_success} shows the intended behavior, where the model speaks at the same moment as the instructor, delivers the same instruction, and stops before the user acts.

The two failure modes we measure follow. In Figure~\ref{fig:case_late} the onset and the content are both right, but the model speaks for 5.8 and 4.0 seconds where the instructor took 1.6 and 1.4, so the user is already acting while the assistant is still talking. This is the failure behind the 28.2\% late-ending rate. In Figure~\ref{fig:case_step} the model guides the wrong step, delivering an instruction that belongs 14 and 17 seconds earlier in the task. These are the pairs for which a neighbouring reference is closer than the matched one.

A third failure comes from the backbone rather than from intervention behavior. In 47 of 1{,}090 utterances the model does not stop, producing degenerate output such as \textit{``Or this, this, put it on the table. Put it on the table. Put it on''} over 21 seconds. Language switching appears in 13 utterances, including a 373-second Chinese repetition loop. These artifacts are inherited from MiniCPM-o 4.5 and are discussed in the Limitations section.

\subsection{Artifacts and Licenses}
\label{app:licenses}

Table~\ref{tab:artifacts} lists all datasets, models, and tools used in this work with their licenses. We use all artifacts for research purposes consistent with their intended use. Our dataset is derived from HoloAssist, which is released under CDLAv2, a permissive license that allows redistribution and derivative works. The original HoloAssist recordings anonymize speaker identities through pitch shifting. Our speech resynthesis replaces all original voices with zero-shot TTS output using unrelated reference speakers from Multilingual LibriSpeech, so no original speaker identity is recoverable from our released data.

\subsection{AI Assistant Usage}
\label{app:ai}

We use \textit{gpt-5.4-mini} for transcript normalization (Section~\ref{sec:construction}) and LLM-as-a-judge evaluation (Appendix~\ref{app:eval}), \textit{gpt-5.5} for action-linked utterance classification, and \textit{gpt-5.6-luna} for utterance type classification (Appendix~\ref{app:eval}). We additionally use an AI assistant for proofreading the manuscript and the human evaluation survey template, for drafting analysis and plotting code, and for generating the initial draft of Figure~\ref{fig:1_overview}.

\end{document}